\documentclass[runningheads]{llncs}

\usepackage{eccv}

\usepackage{eccvabbrv}

\usepackage{graphicx}
\usepackage{booktabs}

\usepackage[accsupp]{axessibility}  

\usepackage{hyperref}

\usepackage{orcidlink}

\usepackage{amsmath}
\usepackage{multirow}
\usepackage{pifont}
\usepackage[dvipsnames]{xcolor}
\DeclareMathOperator*{\argmin}{arg\,min}
\usepackage{booktabs}
\usepackage{textcomp}
\usepackage{caption}
\usepackage{makecell}

\usepackage{algorithm}
\usepackage{algpseudocode}

\begin{document}

\title{Wid3R: Wide Field-of-View 3D Reconstruction via Camera Model Conditioning} 

\titlerunning{Wid3R}

\author{Dongki Jung\inst{1,2} \and  Jaehoon Choi\inst{1} \and Adil Qureshi\inst{1} \and Somi Jeong\inst{2} \and\\ Dinesh Manocha\inst{1} \and Suyong Yeon\inst{2}
}

\authorrunning{Jung et al.}

\institute{$^{1}$University of Maryland, College Park and $^{2}$NAVER LABS}

{
\maketitle
\vspace{-1em}

\begin{figure}[ht]
  \centering
  \includegraphics[width=\textwidth]{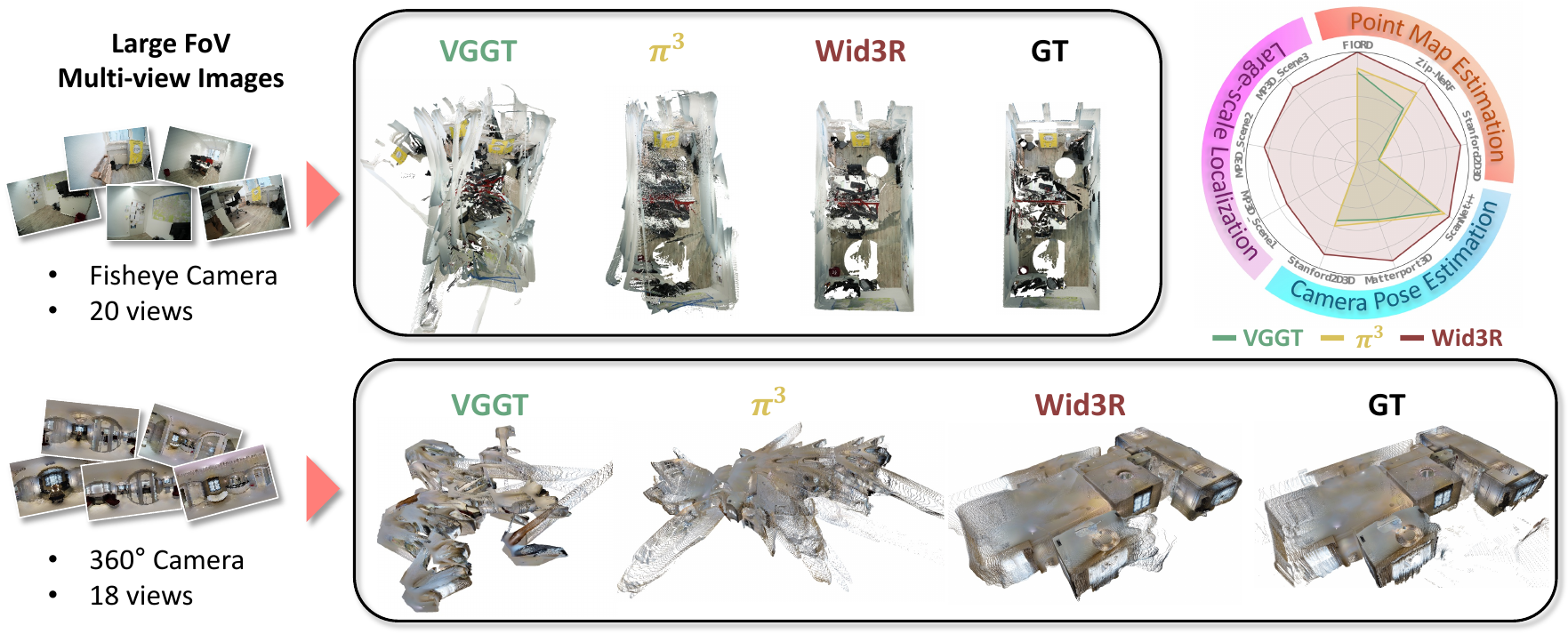}
  \vspace*{-6mm}
  \captionof{figure}{
  %
  Wid3R surpasses VGGT \cite{wang2025vggt} and $\pi^3$ \cite{wang2025pi} in feed-forward reconstruction of fisheye and 360$^\circ$ imagery, demonstrating strong robustness to camera distortion. 
  It achieves superior performance not only in pose and point map estimation but also in large-scale localization and mapping, enabled by its ray-based modules and camera model conditioning.
  }
  \label{fig_teaser}
  \vspace*{-14mm}
\end{figure}

}

\begin{abstract}
    We present Wid3R, a feed-forward neural network for multi-view visual geometry reconstruction that supports wide field-of-view camera models. 
    Unlike existing methods that assume rectified or pinhole inputs, Wid3R directly models wide-angle imagery without explicit calibration or undistortion. 
    Our approach leverages a ray-based representation with spherical harmonics and introduces a novel camera model token to enable distortion-aware reconstruction.
    To the best of our knowledge, Wid3R is the first multi-frame feed-forward 3D reconstruction method that supports 360$^\circ$ imagery. 
    Moreover, we show that conditioning on diverse camera types improves generalization to 360$^\circ$ scenes and alleviates data sparsity issues.
    Wid3R achieves significant performance gains, improving AUC@30$^\circ$ by up to +33.67 on Zip-NeRF (fisheye) and +77.33 on Stanford2D3D (360). Project Page: \href{https://jdk9405.github.io/Wid3R/}{https://jdk9405.github.io/Wid3R/}
\end{abstract}
\section{Introduction}
Modern computer vision systems increasingly rely on accurate 3D geometric understanding across a wide variety of imaging devices. 
While most learning-based multi-view geometry methods \cite{wang2024dust3r,leroy2024grounding,cabon2025must3r,wang2025vggt,wang2025pi} have achieved impressive progress, they overwhelmingly assume that input images are captured with pinhole cameras or have been rectified through calibration and undistortion. 
These assumptions are deeply embedded in both the architectural design of these models and the datasets on which they are trained. 
As a result, existing approaches are inherently biased toward perspective images and struggle to generalize when handling distorted images from wide field-of-view cameras, as shown in Fig. \ref{fig_teaser}.

However, real-world scenarios rarely adhere to these limitations, which often involve wide-angle cameras.
Robotics \cite{courbon2007generic}, autonomous navigation \cite{kumar2023surround}, AR/VR \cite{engel2023project}, and large-scale mapping systems \cite{jung2025edm,jung2025im360,jung2025rpg360} frequently employ fisheye \cite{yeshwanth2023scannet++, duckworth2023smerf} or spherical cameras \cite{chang2017matterport3d,armeni2017joint} to maximize coverage and minimize blind spots. 
These cameras introduce complex non-linear projection effects, and their imaging geometry deviates significantly from the pinhole model.
Consequently, current learning-based techniques \cite{wang2024dust3r, wang2025vggt, wang2025moge, hu2024metric3d}, which are trained solely on pinhole images, often fail to produce reliable depth, pose, or 3D reconstruction outputs in such settings without substantial preprocessing, including manual calibration and rectification.
These preprocessing steps not only introduce engineering overhead but can also degrade geometric fidelity due to information loss during rectification \cite{kumar2020unrectdepthnet,luan2025lifting,piccinelli2025unik3d}.

More recently, researchers have explored learning-based geometry estimation for large field-of-view imagery, either by projecting different camera models into a common equirectangular space \cite{guo2025depth} or by adopting ray-based formulations that implicitly capture projection effects within the network \cite{piccinelli2025unik3d}. 
However, these methods primarily focus on single-view depth or point map estimation and do not directly address multi-view 3D reconstruction. 
To the best of our knowledge, existing work does not provide a generalizable solution for multi-view 3D geometric estimation across cameras with wide fields of view.

\begin{table}[t]
\centering
\caption{
We present Wid3R, the first multi-frame feed-forward network capable of handling distorted images. 
Wid3R enables accurate 3D point map and camera pose estimation, even in the presence of image distortion.
}
\vspace*{-4mm}
\label{tab:model_comparison}
\resizebox{0.75\linewidth}{!}
{
\begin{tabular}{l
@{\hspace{6pt}} c
@{\hspace{10pt}} c
@{\hspace{10pt}} c
@{\hspace{10pt}} c}
\toprule
\multirow{2}{*}{Method} & Input & \multicolumn{1}{c}{Model} & \multicolumn{2}{c}{Output} 
\\
\cmidrule(lr){2-2}\cmidrule(lr){3-3}\cmidrule(lr){4-5}
 & Number of Frames & Distortion Awareness & 3D Point & Pose 
\\
\midrule
DAC \cite{guo2025depth} 
& Single & \textcolor{ForestGreen}{\ding{51}} & \textcolor{ForestGreen}{\ding{51}} & \textcolor{Red}{\ding{55}} 
\\
Unik3D \cite{piccinelli2025unik3d} 
& Single & \textcolor{ForestGreen}{\ding{51}} & \textcolor{ForestGreen}{\ding{51}} & \textcolor{Red}{\ding{55}}
\\ 
Dust3R~\cite{wang2024dust3r} 
& Pair & \textcolor{Red}{\ding{55}} & \textcolor{ForestGreen}{\ding{51}} & \textcolor{ForestGreen}{\ding{51}}
\\
VGGT~\cite{wang2025vggt} 
& Multi & \textcolor{Red}{\ding{55}} & \textcolor{ForestGreen}{\ding{51}} & \textcolor{ForestGreen}{\ding{51}}
\\

$\pi^3$~\cite{wang2025pi} 
& Multi & \textcolor{Red}{\ding{55}} & \textcolor{ForestGreen}{\ding{51}} & \textcolor{ForestGreen}{\ding{51}}
\\
\textbf{Wid3R} \textbf{(Ours)}
& Multi & \textcolor{ForestGreen}{\ding{51}} & \textcolor{ForestGreen}{\ding{51}} & \textcolor{ForestGreen}{\ding{51}}
\\
\bottomrule
\end{tabular}
}
\vspace*{-4mm}
\end{table}

\noindent\textbf{Main Results}\quad
As shown in Table \ref{tab:model_comparison}, we present Wid3R, the first feed-forward neural network for multi-view 3D reconstruction from wide field-of-view images captured with fisheye or spherical cameras. Given a set of distorted multi-view images, Wid3R jointly estimates dense 3D point maps and camera poses without requiring explicit rectification or camera-specific preprocessing. Wid3R utilizes a ray-based geometric representation that models each pixel as a camera ray and encodes ray directions using spherical harmonics \cite{piccinelli2025unik3d}. This formulation enables the network to reason directly about scene geometry across diverse projection models and distortion characteristics within a unified network.
In addition, we introduce a camera model token as a lightweight conditioning mechanism that enables the network to adapt to different camera models.
Together, these components establish Wid3R as a multi-view geometric foundation model that operates in ray space with camera-aware conditioning, enabling robust 3D reconstruction across diverse datasets captured by wide-angle cameras \cite{duckworth2023smerf,gunes2025fiord,armeni2017joint,yeshwanth2023scannet++,chang2017matterport3d}.

In summary, the novel contributions of our work include:
\begin{itemize}
    \item We introduce the first multi-view feed-forward network that operates directly on distorted images captured with wide field-of-view cameras.
    \item We propose a unified ray-based representation for multi-view 3D reconstruction that encodes camera geometry with spherical harmonics, enabling the model to robustly handle distortion.
    \item We design camera model tokens as an explicit conditioning mechanism that informs the network of projection characteristics specific to each camera model within a shared architecture.
\end{itemize}
We validate Wid3R through comprehensive evaluations across multiple datasets, achieving up to $3.38\%$ improvement in $\delta_{1.25}$ for depth estimation on Matterport3D \cite{chang2017matterport3d}, $82.23\%$ improvement in AUC$@30^\circ$ for camera pose estimation on Zip-NeRF \cite{duckworth2023smerf}, and $48.2\%$ accuracy improvement in point map reconstruction on Stanford2D3D \cite{armeni2017joint}.
\section{Related Work}
\subsection{Conventional 3D Reconstruction}
Structure from Motion (SfM) \cite{schonberger2016structure} is a fundamental problem in computer vision that aims to recover camera poses and 3D structure from a collection of images.
Traditional SfM pipelines consist of multiple sequential stages. These typically begin with keypoint detection and descriptor extraction \cite{SIFT,detone2018superpoint,sarlin2020superglue,revaud2019r2d2}, followed by feature matching across images \cite{sarlin2020superglue}. 
Based on the established correspondences, two-view geometry is estimated using robust methods such as RANSAC \cite{RANSAC} to recover relative camera poses \cite{hartley2003multiple}. 
SfM systems then proceed with incremental triangulation, where new images are registered and additional 3D points are reconstructed. 
The estimated camera parameters and 3D structure are further refined through bundle adjustment \cite{Bundleadjustment}. 
The resulting sparse reconstruction can subsequently be used as input for multi-view stereo \cite{furukawa2015multi,schonberger2016pixelwise} or neural rendering methods \cite{mildenhall2021nerf,kerbl20233d} to obtain dense scene representations.
Several works have proposed improvements to this classical SfM pipeline. 
GLOMAP \cite{GLOMAP} adopts a global optimization strategy that estimates geometry for all input images simultaneously, improving robustness to local errors. 
DetectorFreeSfM \cite{he2024detector} shows that detector-free or dense matching approaches \cite{sun2021loftr,edstedt2023dkm,edstedt2024roma} can significantly improve reconstruction quality in texture-poor environments where keypoint-based methods struggle. 
While most prior work assumes rectified or undistorted pinhole images, IM360 \cite{jung2025im360} shows that spherical cameras can be effective for sparse indoor reconstruction. 
By leveraging detector-free matching \cite{jung2025edm} on spherical imagery, this approach extends SfM to wide field-of-view camera models.
Despite these advances, existing SfM methods continue to rely on multi-stage pipelines that can introduce computational bottlenecks, such as iterative optimization and dense matching across large numbers of image pairs.
\subsection{Feed-Forward 3D Reconstruction}
Recent advances in feed-forward reconstruction models have shown that scene-level 3D structure can be estimated directly from image inputs without relying on iterative optimization at test time. 
DPT \cite{ranftl2021vision} demonstrates that Vision Transformers \cite{dosovitskiy2020image} can significantly improve dense prediction quality for monocular depth estimation by leveraging global image context. 
Metric3D \cite{yin2023metric3d} introduces a canonical camera transformation module that enables monocular metric depth estimation by normalizing camera geometry, while MoGe \cite{wang2025moge} proposes affine-invariant point map estimation coupled with optimal alignment to recover consistent geometric structure.

To address consistency across views, DUSt3R \cite{wang2024dust3r} formulates feed-forward pairwise reconstruction, where an image pair is jointly processed to predict dense point clouds expressed in the coordinate frame of one camera. 
Building on this formulation, Spann3R \cite{wang20243d} incorporates an external spatial memory to support incremental reconstruction over multiple views, while FLARE \cite{zhang2025flare} modifies the architecture to progressively predict 3D points under the estimated camera pose.
VGGT \cite{wang2025vggt} further improves robustness and accuracy by leveraging large-scale training data and multi-task supervision to estimate geometry in a global world reference frame. 
$\pi^3$ \cite{wang2025pi} addresses sensitivity to reference view selection by introducing a permutation-equivariant architecture that reduces performance variance across different view orderings.
%
%
%
Recently, Depth Anything 3 \cite{lin2025depth} and MapAnything \cite{keetha2025mapanything} utilized ray maps to represent camera parameters.
However, these feed-forward reconstruction methods still assume distortion-free perspective images and pinhole camera models, resulting in degraded performance on wide field-of-view imagery.
We overcome this limitation by operating directly in ray space with camera-aware conditioning, enabling robust multi-view reconstruction in the presence of lens distortion.

\subsection{Wide Field-of-View Camera Models}
Camera calibration plays a critical role in recovering 3D structure from the 2D image plane by estimating intrinsic parameters such as focal length, principal point, and lens distortion. 
A variety of camera models have been proposed to capture different imaging geometries, including the pinhole model, the Kannala–Brandt model \cite{kannala2006generic}, the Mei model \cite{mei2007single}, the Unified Camera Model \cite{geyer2000unifying,khomutenko2015enhanced}, the Double Sphere model \cite{usenko2018double}, and omnidirectional camera models \cite{scaramuzza2021omnidirectional}. 
However, approaches based on explicit camera modeling are sensitive to data acquisition conditions, often leading to variability in the estimated calibration parameters \cite{abbas2019analysis,pollefeys2000some}. 
Moreover, as camera models become more complex with increased parameterization, the calibration process may introduce additional sources of error and instability \cite{clarke1998development,lochman2021babelcalib}.

As an alternative to explicit camera modeling, learning-based  methods have emerged that perform monocular point estimation by representing images as pencils of rays in a spherical formulation \cite{piccinelli2024unidepth,piccinelli2025unik3d}. 
In particular, UniK3D \cite{piccinelli2025unik3d} removes explicit camera assumptions by modeling scene geometry directly in ray space, leveraging a spherical harmonics basis. 
Recent work adapts pretrained monocular depth estimation models to fisheye cameras by introducing additional calibration tokens \cite{gangopadhyay2025extending}.
In contrast, we encode camera models through dedicated tokens within a ray-based geometric representation and extend this formulation to multi-view reconstruction, enabling adaptation to wide-angle camera types, including fisheye and 360$^\circ$ imagery.
\section{Our Approach: Wid3R}
Generalizable 3D geometry estimation across diverse camera configurations is essential for real-world applications. We introduce a novel method that enables multi-view 3D reconstruction from wide field-of-view camera images. We first present the problem formulation in Section~\ref{sec:method:definition}, followed by a discussion of the local coordinate representation in Section~\ref{sec:method:local_geometry_representation}, and finally describe the camera configuration modeled through a ray-based representation in Section~\ref{sec:method:camera}.

%
\subsection{Problem Definition}
\label{sec:method:definition}
Given a sequence of $N$ RGB images $\{I_i\}_{i=1}^{N}$, where each image $I_i \in \mathbb{R}^{3 \times H \times W}$, the Wid3R network $f(\cdot)$ maps the multi-view image sequence into a unified 3D geometric representation:
\begin{equation}
\begin{aligned}
    f\big(\{\mathbf{I}_{i}, \mathbf{C}_{i}\}_{i=1}^N\big)
    = \{\mathbf{T}_{i}, \boldsymbol{\mathcal{R}}_{i}, \mathbf{D}_{i}, \mathbf{U}_{i}\}_{i=1}^{N},
\end{aligned}
\label{eq:basic}
\end{equation}
where $\mathbf{C}_i$ denotes a camera model token associated with view $i$.
$\mathbf{T}_i \in \mathbb{R}^{4 \times 4}$ represents the camera pose,
$\boldsymbol{\mathcal{R}}_i \in \mathbb{R}^{3 \times H \times W}$ denotes the pencil of rays defined at each image location,
$\mathbf{D}_i \in \mathbb{R}^{H \times W}$ corresponds to the radial distance along each ray, and
$\mathbf{U}_i \in \mathbb{R}^{H \times W}$ represents the associated uncertainty.
Here, we incorporate the camera model token as an input condition for geometric prediction, enabling robust reconstruction from wide field-of-view images without explicit calibration.
\vspace*{-4mm}
%
%
%
%
\begin{figure*}[t]
  \centering
  \includegraphics[width=0.7\textwidth]{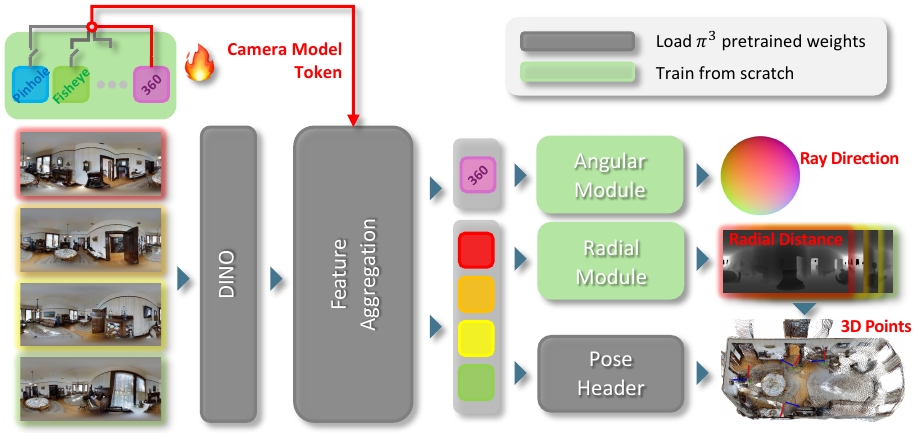}
  \vspace*{-2mm}
  \caption{
  Overview of the architecture. 
  Multi-view wide field-of-view images are processed by DINO and aggregated through a feature aggregation module. 
  An appropriate camera model token is selected to condition the network on camera-specific priors. 
  Instead of directly regressing point or depth maps, Wid3R decomposes 3D reconstruction into angular and radial components, predicting ray directions and radial distances that are robust to camera projection distortion. 
  A pose header estimates camera poses, enabling the reconstruction of global 3D points. 
  Pretrained $\pi^{3}$ \cite{wang2025pi} weights are loaded where applicable to accelerate training convergence, while the remaining components are trained from scratch.
  }
  \label{fig_network}
  \vspace*{-4mm}
\end{figure*}
\subsection{Per-View Local Geometry Reconstruction}
\label{sec:method:local_geometry_representation}
Wide field-of-view images provide extensive scene coverage, enabling large portions of a scene to be captured with a relatively small number of images. 
For example, 360$^\circ$ cameras observe the entire surrounding environment at each viewpoint, which often leads to sparsely distributed scene captures along a trajectory \cite{jung2025im360}. 
In such settings, consecutive views may exhibit limited visual overlap, particularly in long-term sequences or large-scale environments. 
Under such conditions, methods that rely on view-dependent assumptions \cite{wang2025vggt} may struggle when reference and source views lack sufficient overlap.

To address this challenge, we adopt a permutation-equivariant network design inspired by \cite{wang2025pi}, in which scene geometry is represented in per-view local coordinate frames rather than a shared global reference frame.
%
For each input image $\mathbf{I}_{i}$, our network estimates a pixel-aligned 3D point map $\hat{\mathbf{P}}_{i} = \hat{\boldsymbol{\mathcal{R}}}_i \odot \hat{\mathbf{D}}_i$, obtained via element-wise multiplication between the ray directions $\hat{\boldsymbol{\mathcal{R}}}_i$ and the corresponding radial distances $\hat{\mathbf{D}}_i$.
Since each point map is expressed in its own local camera coordinate system, scale ambiguity arises across the $N$ images.
Some prior works in monocular depth estimation address this issue by estimating scale differences either with respect to the ground truth \cite{wang2025moge} or across the predicted point maps \cite{jung2025more}. 
Rather than resolving scale on a per-image basis, $\pi^3$ \cite{wang2025pi} assumes that all predicted point maps share a single, unknown but consistent scale factor across the multi-view images.
We follow this scaling and align the predicted point maps $\{\mathbf{\hat{P}}_i\}_{i=1}^{N}$ to the corresponding ground-truth set $\{\mathbf{P}_i\}_{i=1}^{N}$ by solving for a single optimal scale factor $s^{*}$ \cite{wang2025moge} as follows,
\begin{equation}
\begin{aligned}
    s^{*} = \argmin_{s} {\sum_{i=1}^{N}\sum_{j=1}^{HW} \frac{1}{\mathbf{D}_{i, j}}\| s\hat{\mathbf{P}}_{i, j} - \mathbf{P}_{i, j} \|_1},
\end{aligned}\label{eq:scale}
\end{equation}
where $\hat{\mathbf{P}}_{i,j}, \mathbf{P}_{i,j} \in \mathbb{R}^3$ denote the predicted and ground-truth 3D points at index $j$ of the point maps $\hat{\mathbf{P}}_{i}$ and $\mathbf{P}_{i}$, respectively.
Here, instead of using perpendicular depth as the weight in the $\ell_1$ loss, we utilize radial distance $\mathbf{D}_{i,j}$ to accommodate arbitrary camera models.
The resulting weighted $\ell_1$ loss over the entire image sequence ensures consistent scale alignment across all views.
Using this optimal scale term, the point map loss is defined as:
\begin{equation}
\begin{aligned}
    \mathcal{L}_{\text{points}} = \frac{1}{3NHW} \sum_{i=1}^{N} \sum_{j=1}^{HW} \frac{1}{\mathbf{D}_{i,j}} \| s^{*}\hat{\mathbf{P}}_{i,j} - \mathbf{P}_{i,j} \|_1.
\end{aligned}\label{eq:loss_points}
\end{equation}
We incorporate a normal loss to enforce local surface smoothness by minimizing the angular difference between predicted normals $\hat{\mathbf{n}}_{i,j}$ and ground truth normals $\mathbf{n}_{i,j}$.
Normals are computed using an 8-neighbor convention on the point map \cite{yang2018unsupervised,jung2021dnd}, by taking the cross products of vectors formed with neighboring points on the image grid,
\begin{equation}
\begin{aligned}
    \mathcal{L}_\text{normal} = \sum_{i=1}^{N} \sum_{j=1}^{HW} \text{arccos}(\hat{\mathbf{n}}_{i,j} \cdot \mathbf{n}_{i,j}).
\end{aligned}\label{eq:loss_normal}
\end{equation}
%
%
Our network predicts the camera pose $\hat{\mathbf{T}}_i$ corresponding to each image $\mathbf{I}_{i}$
The relative pose is computed between the different frames:
\begin{equation}
\begin{aligned}
    \hat{\mathbf{T}}_{i \leftarrow j} = \hat{\mathbf{T}}_{i}^{-1} \hat{\mathbf{T}}_{j},
\end{aligned}\label{eq:rel_pose}
\end{equation}
where $\hat{\mathbf{T}}_{i \leftarrow j}$ consists of a rotation $\hat{\mathbf{R}}_{i \leftarrow j} \in SO(3)$ and a translation $\hat{\mathbf{t}}_{i \leftarrow j} \in \mathbb{R}^{3}$.
The camera pose loss is defined as a weighted average of a rotation loss term and a translation loss term:
\begin{equation}
\begin{aligned}
    \mathcal{L}_{\text{pose}} = \frac{1}{N(N-1)} \sum_{i \neq j} \big(\mathcal{L}_{\text{rot}}(i, j) + \lambda_{\text{trans}} \mathcal{L}_{\text{trans}}(i, j)\big).
\end{aligned}\label{eq:loss_pose}
\end{equation}
The rotation loss minimizes the geodesic distance between the predicted relative rotation $\hat{\mathbf{R}}_{i \leftarrow j}$ and its ground truth counterpart $\mathbf{R}_{i \leftarrow j}$.
To align the predicted 3D point maps $\{\hat{\mathbf{P}}_{i}\}_{i=1}^{N}$ and camera poses $\{\hat{\mathbf{T}}_{i}\}_{i=1}^{N}$ in a shared global coordinate, 
we apply the optimal scale factor $s^{*}$ to the predicted relative camera translations $\hat{\mathbf{t}}_{i \leftarrow j}$ so that they are aligned with the ground truth $\mathbf{t}_{i \leftarrow j}$.
\begin{equation}
\left\{
\begin{aligned}
    &\mathcal{L}_{\text{rot}}(i, j) = \text{arccos}\left(\frac{\text{Tr}\big((\mathbf{R}_{i \leftarrow j})^{\top} \hat{\mathbf{R}}_{i \leftarrow j}\big) - 1}{2}\right)
    \\
    &\mathcal{L}_{\text{trans}}(i, j) = \ell_{\delta}(s^{*}\hat{\mathbf{t}}_{i \leftarrow j} - \mathbf{t}_{i \leftarrow j})
\end{aligned}\label{eq:loss_trans}
\right.
\end{equation}
where $\ell_\sigma(\cdot)$ denotes the Huber loss, which is used to reduce the influence of outliers.


\subsection{Ray-Based Feed-Forward 3D Reconstruction}
\label{sec:method:camera}
All existing feed-forward 3D reconstruction methods \cite{wang2025vggt,wang2025pi} adopt point maps as their output representation, and accordingly employ a dedicated point-map head in the network architecture. 
While effective for standard perspective cameras, this design choice limits the ability to generalize across diverse camera models, especially those with wide fields of view.
To address this limitation, we introduce a multi-frame feed-forward 3D reconstruction method tailored for wide field-of-view imagery. 
Building on UniK3D \cite{piccinelli2025unik3d}, we represent different camera models using spherical harmonics (SH) within a pencil-of-rays formulation, enabling a compact and accurate angular parameterization well suited for wide field-of-view imagery.
This formulation establishes a bijective mapping between the polar angle $\theta$ and azimuthal angle $\phi$ in spherical coordinates and 3D Cartesian directions. 
By leveraging a predefined SH basis $Y^{lm}(\cdot, \cdot)$, arbitrary camera geometries can be implicitly approximated using a compact set of SH coefficients.
Inspired by this representation, we replace the conventional point-map head with two decoupled components: an Angular Module and a Radial Module.
Furthermore, we introduce a novel trainable camera token $\mathbf{C}_{i}$, which acts as a prompt to explicitly differentiate camera models while being implicitly integrated into the network architecture. The Angular Module predicts SH coefficients $\hat{\boldsymbol{k}}^{lm}_{i}(\cdot)$ conditioned on the camera token $\mathbf{C}_{i}$. Given these coefficients and the basis, the pencil of rays $\hat{\boldsymbol{\mathcal{R}}}_{i}(\cdot,\cdot)$ is computed as:
\begin{equation}
\begin{aligned}
    \hat{\boldsymbol{\mathcal{R}}}_{i}(\theta, \phi) = \sum_{l=0}^{L} \sum_{m=-l}^{l} \hat{\boldsymbol{k}}^{lm}_{i}(\mathbf{C}_{i}) \ {Y}^{lm}(\theta, \phi),
\end{aligned}\label{eq:ray}
\end{equation}
where $l$ and $m$ denote the degree and order of the harmonics, respectively.
The Radial Module then estimates the radial distance $\hat{\mathbf{D}}_{i}$ and its associated uncertainty $\hat{\mathbf{U}}_{i}$ along each ray $\hat{\boldsymbol{\mathcal{R}}}_{i}$.
Using these two modules, the predicted 3D point map can be expressed as $\hat{\mathbf{P}}_{i} = \hat{\boldsymbol{\mathcal{R}}}_{i} \odot \hat{\mathbf{D}}_{i}$ with these separate modules.
%
%
For the ray loss, to compensate for the limited number of wide field-of-view images during training, we adopt an asymmetric formulation following \cite{piccinelli2025unik3d},
\begin{equation}
\begin{aligned}
    \mathcal{L}_\text{ray} &= \beta \mathcal{L}_\theta^{0.7} + (1-\beta) \mathcal{L}_{\phi}^{0.5},
    \\
    \text{with}\quad
    \mathcal{L}_{\omega}^{\alpha} = \alpha \sum_{\hat{\omega} > \omega^{*}} &\left| \hat{\omega} - \omega \right| + (1-\alpha)\sum_{\hat{\omega} \leq \omega} \left| \hat{\omega} - \omega^{*} \right|,
\end{aligned}
\label{eq:loss_ray}
\end{equation}
where $\omega \in \{\theta, \phi\}$ and $\beta = 0.75$.
The radial loss is defined as the L1 loss between the predicted radial distance $\hat{\mathbf{D}}_{i}$ and the ground-truth distance $\mathbf{D}_{i}$.
The uncertainty loss is the L1 loss between the radial loss and the predicted uncertainty,
%
\begin{equation}
\begin{aligned}
    \mathcal{L}_{\text{rad}} = \| \hat{\mathbf{D}}_{i} - \mathbf{D}_{i} \|,
\end{aligned}\label{eq:loss_rad}
\end{equation}
\vspace*{-2mm}
\begin{equation}
\begin{aligned}
    \mathcal{L}_{\text{uncer}} = \left\| |\hat{\mathbf{D}}_{i} - \mathbf{D}_{i}| - \hat{\mathbf{U}}_{i} \right\|.
\end{aligned}\label{eq:loss_uncert}
\end{equation}
%


\begin{figure}[t]
  \centering
  \begin{minipage}[c]{0.54\linewidth}
    \includegraphics[width=\linewidth]{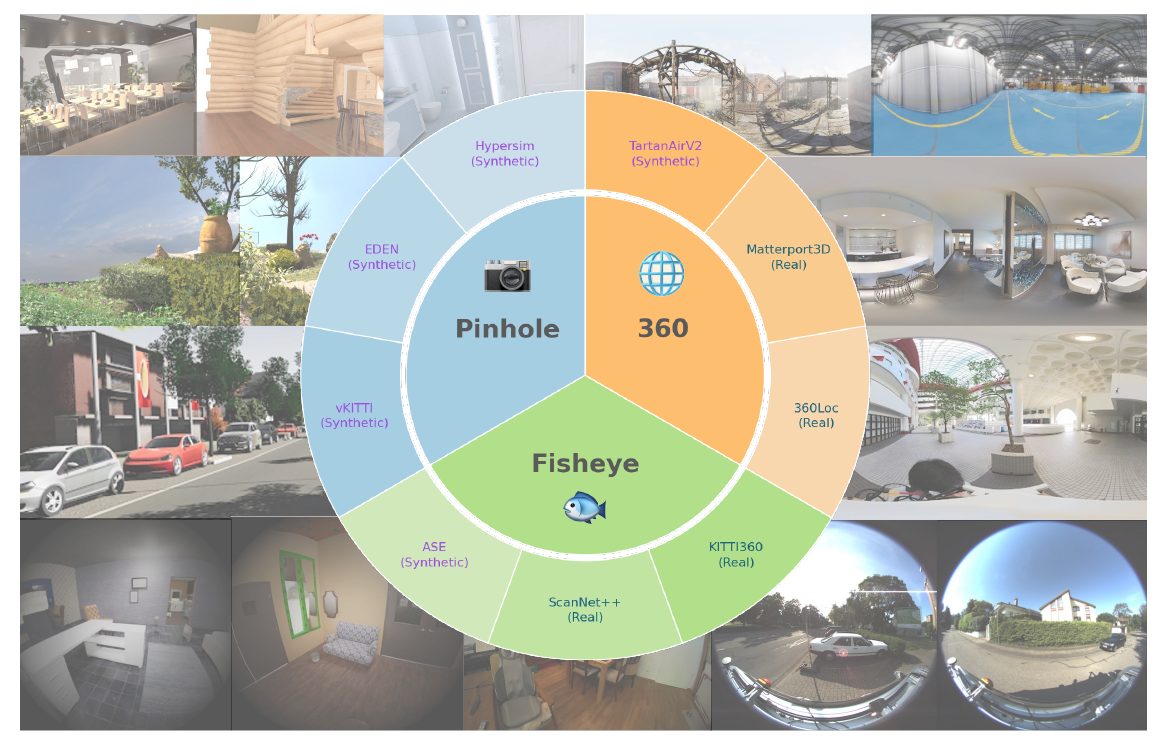}
  \end{minipage}\hfill
  \begin{minipage}[c]{0.45\linewidth}
    \caption{
    Composition of camera models and training datasets. 
    The diagram illustrates diverse camera model configurations, including pinhole, fisheye, and 360$^\circ$ cameras, along with representative scenes. 
    Wid3R unifies these camera models within a single network, enabling robust reconstruction under wide field-of-view settings.
    }
    \label{fig_data}
  \end{minipage}
\end{figure}
\subsection{Training}\label{sec:Training}
\noindent\textbf{Training Losses}
All loss terms are combined using a weighted sum, and the model is trained end-to-end by minimizing the total loss function.
We set $\lambda_{\text{normal}} = 10$, $\lambda_{\text{pose}} = 0.1$, $\lambda_{\text{ray}} = 1.0$, $\lambda_{\text{rad}} = 1.0$, and $\lambda_{\text{uncer}} = 0.1$ in all experiments,
\begin{equation}
\begin{aligned}
    \mathcal{L}_{\text{total}}\quad = \quad&\mathcal{L}_{\text{points}} \ \ + \ \ \lambda_{\text{normal}}\mathcal{L}_{\text{normal}}\ \ + \ \ \lambda_{\text{pose}}\mathcal{L}_{\text{pose}} 
    \\
    +\ \ &\lambda_{\text{ray}}\mathcal{L}_{\text{ray}} \ \ +\ \ \lambda_{\text{rad}}\mathcal{L}_{\text{rad}} \ \ + \ \ \lambda_{\text{uncer}}\mathcal{L}_{\text{uncer}}.
\end{aligned}\label{eq:loss_total}
\end{equation}
%

\noindent\textbf{Training Datasets}
We train our model using a large and diverse collection of datasets that span a wide field-of-view camera configurations, including TartanAirV2 \cite{wang2020tartanair}, ASE \cite{engel2023project}, Hypersim \cite{roberts2021hypersim}, KITTI-360 \cite{liao2022kitti}, 360Loc \cite{huang2024360loc}, Matterport3D \cite{chang2017matterport3d}, ScanNet++ \cite{yeshwanth2023scannet++}, EDEN \cite{le2021eden}, and VKITTI \cite{cabon2020virtual}. 
Collectively, these datasets cover three camera categories (pinhole, fisheye, and 360$^\circ$ cameras), with multiple datasets available for each configuration, enabling training across diverse camera projection models, as shown in Fig.~\ref{fig_data}. 
Among these datasets, five are synthetic and four are captured from real-world environments, providing a balance between large-scale, noise-free supervision and realistic scene complexity. 

Since most of the training data consists of large-scale scenes, it is important to sample input images within each batch such that sufficient visual overlap is maintained. 
To this end, we compute pairwise distances between all images based on their camera positions, forming a 2D distance matrix. 
We then apply a softmax operation to the negative distances to obtain a probability matrix, which is used to sample input images with higher probability for closer viewpoints (see the Supplementary Material for details).
Notably, training data for 360$^\circ$ cameras is significantly more limited than that for other camera types, accounting for less than $1\%$ of the total number of images. 
To better understand this imbalance, we further analyze how incorporating data from different camera models influences performance under 360$^\circ$ camera settings. The results of this analysis are reported in Table \ref{table:ablation}.

\noindent\textbf{Augmentations}
During training, we randomly apply a set of data augmentations that include both geometric and appearance transformations, such as random resizing and cropping, horizontal flipping, and photometric adjustments of brightness, contrast, saturation, gamma, and hue.
In addition, to increase data diversity for wide field-of-view images, we adopt a camera augmentation strategy \cite{piccinelli2025unik3d} that transforms pinhole camera images into fisheye camera models. 
Specifically, we first unproject the 2D depth map obtained under a pinhole camera model into a 3D point cloud, and then reproject the points onto the 2D image plane of a randomly sampled distorted camera model. 
To preserve fine details during this process, we also apply softmax-based splatting \cite{niklaus2020softmax} to warp both images and depth maps.
For equirectangular projection (ERP) images, we further exploit the fact that a single ERP image represents the full 360$^\circ$ viewing sphere \cite{jung2025edm}. 
We apply rotation-based augmentation by randomly sampling the azimuth angle $\in [0, 2\pi]$ and the elevation angle $\in [0, \pi]$, and rotating the image, depth and pose accordingly.
To maintain geometric consistency, the same augmentations are applied consistently to all multi-view input images from the same scene within a batch.

\section{Implementation and Performance}
\subsection{Experiment Settings}

\noindent\textbf{Implementation Details}
The networks are optimized for geometric reconstruction using the Adam optimizer \cite{kingma2014adam} on RTX A6000 or H100 GPUs, starting with an initial learning rate of $5 \times 10^{-5}$
and applying exponential decay at each iteration. Input images are resized to $518 \times 336$.
We initialize the $\pi^{3}$ encoder, feature aggregation module, and pose header from pretrained weights and train all components end-to-end for faster convergence.


\begin{table*}[t]
    \begin{center}
    \caption{\label{table:pose_angle} 
    Zero-shot pose estimation on wide field-of-view images.
    Metrics report the ratio of samples with rotation and translation errors within 30$^\circ$ (higher is better).
    Ours denotes models trained with our composed wide-angle training data, implemented on the $\pi^3$ baseline (Ours ($\pi^3$)) and our ray-based modules (Ours (Wid3R)).
    }
    \vspace*{-2mm}
    \resizebox{0.95\textwidth}{!}{
    \begin{tabular}{l ccc ccc ccc}
        \toprule
        \multirow{2}{*}{Method} & \multicolumn{3}{c}{FIORD \cite{gunes2025fiord}} & \multicolumn{3}{c}{Zip-NeRF \cite{duckworth2023smerf}} & \multicolumn{3}{c}{Stanford2D3D \cite{armeni2017joint}} \\
        \cmidrule{2-4}\cmidrule(lr){5-7}\cmidrule(lr){8-10}
        & RRA@30 $\uparrow$ & RTA@30 $\uparrow$ & AUC@30 $\uparrow$ & RRA@30 $\uparrow$ & RTA@30 $\uparrow$ & AUC@30 $\uparrow$ & RRA@30 $\uparrow$ & RTA@30 $\uparrow$ & AUC@30 $\uparrow$ \\
        \midrule
        VGGT \cite{wang2025vggt} & 82.34 & 84.01 & 44.63 & 64.42 & 65.29 & 20.68 & 19.30 & 33.14 & 2.60 \\
        $\pi^3$ \cite{wang2025pi} & 85.48 & 87.42 & 48.35 & 82.65 & 86.66 & 40.94 & 19.94 & 31.99 & 2.06 \\
        Ours ($\pi^3$ \cite{wang2025pi}) & 99.87 & \textbf{96.39} & \textbf{72.56} & 83.04 & 89.48 & 64.44 & 82.91 & 87.83 & 62.14 \\
        Ours (Wid3R) & \textbf{100.0} & 93.38 & 68.20 & \textbf{93.34} & \textbf{95.15} & \textbf{74.61} & \textbf{94.05} & \textbf{93.29} & \textbf{79.93} \\
        \bottomrule
    \end{tabular}
    }
    \end{center}
    \vspace*{-4mm}
\end{table*}

\begin{table*}[t]
    \begin{center}
    \caption{\label{table:pose_distance} 
    Zero-shot pose estimation on wide field-of-view images.
    Metrics measure rotation and translation distance errors (lower is better).
    }
    \vspace*{-2mm}
    \resizebox{0.95\textwidth}{!}{
    \begin{tabular}{l ccc ccc ccc}
        \toprule
        \multirow{2}{*}{Method} & \multicolumn{3}{c}{FIORD \cite{gunes2025fiord}} & \multicolumn{3}{c}{Zip-NeRF \cite{duckworth2023smerf}} & \multicolumn{3}{c}{Stanford2D3D \cite{armeni2017joint}} \\
        \cmidrule{2-4}\cmidrule(lr){5-7}\cmidrule(lr){8-10}
        & ATE $\downarrow$ & RPE trans $\downarrow$ & RPE rot $\downarrow$ & ATE $\downarrow$ & RPE trans $\downarrow$ & RPE rot $\downarrow$ & ATE $\downarrow$ & RPE trans $\downarrow$ & RPE rot $\downarrow$ \\
        \midrule
        VGGT \cite{wang2025vggt} & 0.54 & 0.99 & 15.02 & 1.83 & 1.73 & 34.41 & 3.07 & 5.01 & 98.83 \\        
        $\pi^3$ \cite{wang2025pi} & 0.53 & 0.90 & 13.45 & 0.85 & 0.87 & 18.55 & 2.83 & 5.01 & 91.13 \\
        Ours ($\pi^3$ \cite{wang2025pi}) & \textbf{0.32} & \textbf{0.51} & 3.99 & 0.85 & 0.67 & 12.93 & 1.43 & 2.21 & 32.94 \\
        Ours (Wid3R) & 0.44 & 0.71 & \textbf{3.27} & \textbf{0.49} & \textbf{0.47} & \textbf{7.72} & \textbf{1.11} & \textbf{1.83} & \textbf{18.99} \\
        \bottomrule
    \end{tabular}
    }
    \end{center}
    \vspace*{-4mm}
\end{table*}

\begin{table*}[t]
    \begin{center}
    \caption{\label{table:vl_mp3d} 
    Quantitative evaluation of large-scale localization performance on the Matterport3D dataset \cite{chang2017matterport3d}.
    Our method demonstrates superior registration accuracy and achieves comparable performance in large-scale scene pose estimation with significantly reduced runtime.
    }
    \vspace*{-2mm}
    \resizebox{0.99\textwidth}{!}{
    \begin{tabular}{l| cccc cccc cccc |c}
        \toprule
        \multirow{2}{*}{Method} & \multicolumn{4}{c}{Matterport3D Scene1} & \multicolumn{4}{c}{Matterport3D Scene2} & \multicolumn{4}{c}{Matterport3D Scene3} & \multicolumn{1}{|c}{\multirow{2}{*}{ Time }} \\
        \cmidrule{2-5}\cmidrule(lr){6-9}\cmidrule(lr){10-13}
        & \# Registered & AUC @3\textdegree & AUC @5\textdegree & AUC @10\textdegree & \# Registered & AUC @3\textdegree & AUC @5\textdegree & AUC @10\textdegree & \# Registered & AUC @3\textdegree & AUC @5\textdegree & AUC @10\textdegree & 
        \\
        \midrule
        OpenMVG \cite{moulon2016openmvg} & 5 / 37 & 0.82 & 1.10 & 1.30 &2 / 20 & 0.27&0.37 &0.45 &25 / 31 & 45.11 &52.82 &58.67 & $\sim$ 3min  \\
        SPSG \cite{detone2018superpoint,sarlin2020superglue} $+$ COLMAP \cite{schonberger2016structure} & 16 / 37 & 14.30 & 15.79 & 16.90 & 12 / 20 & 26.64 & 29.88 & 32.31 & \textbf{31 / 31} & 75.02 & 85.01 & 92.51 & $\sim$ 5min  \\
        SphereGlue \cite{gava2023sphereglue} $+$ COLMAP \cite{schonberger2016structure} & 21 / 37 & 23.95 & 26.98 & 29.26 & 12 / 20 & 23.11  & 27.76 & 31.24 & \textbf{31 / 31} & 66.83 & 80.06 & 90.03 & $\sim$ 5min \\
        DKM \cite{edstedt2023dkm} $+$ COLMAP \cite{schonberger2016structure} & 22 / 37& 25.14 & 27.70 & 29.62 & 9 / 20 & 14.64 & 16.36 & 17.66 & \textbf{31 / 31} & \textbf{75.94} & \textbf{85.56} & \textbf{92.78} & $\sim$ 30min  \\
        EDM \cite{jung2025edm} $+$ IM360 \cite{jung2025im360} & \textbf{37 / 37} & \textbf{49.16} & \textbf{69.05} & \textbf{84.53} & \textbf{20 / 20} & 34.91 & 44.16 & 66.87 & \textbf{31 / 31} & 73.95 & 84.37 & 92.18 & $\sim$ 30min \\
        \midrule
        Ours (Wid3R) & \textbf{37 / 37} & 31.64 & 51.02 & 72.81 & \textbf{20 / 20} & \textbf{51.22} & \textbf{69.66} & \textbf{84.77} & \textbf{31 / 31} & 64.74 & 78.77 & 89.40 & \textbf{$\sim$ 3s} \\
        \bottomrule
    \end{tabular}
    }
    \end{center}
    \vspace*{-6mm}
\end{table*}
\begin{figure*}[t]
  \centering
  \includegraphics[width=0.95\textwidth]{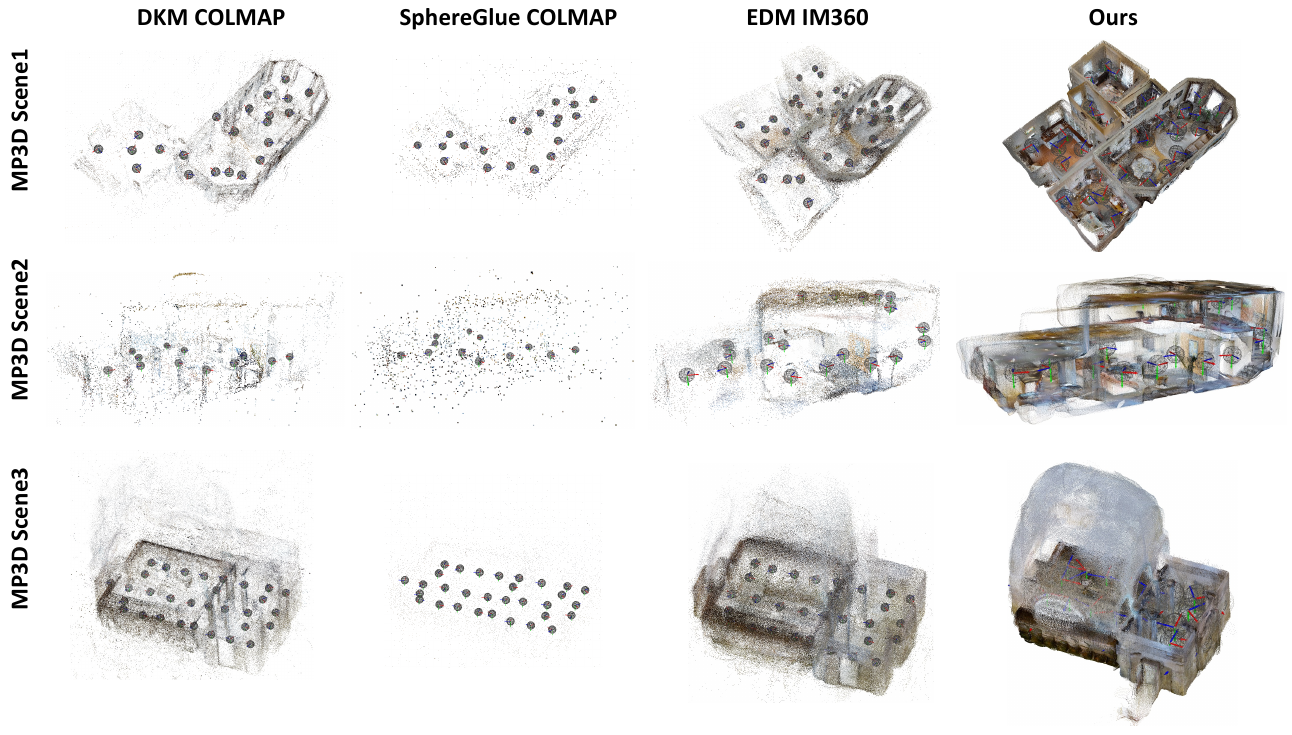}
  \vspace*{-5mm}
  \caption{
  Qualitative results of 3D point reconstruction with visual localization. 
  Previous methods estimate 3D points through triangulation, whereas our method directly predicts them in a feed-forward manner. 
  Our approach produces more complete and consistent reconstructions across large-scale Matterport3D (MP3D) \cite{chang2017matterport3d} scenes.
  }
  \label{fig_sfm}
  \vspace*{-10mm}
\end{figure*}

\begin{figure*}[t]
  \centering
  \includegraphics[width=0.96\textwidth]{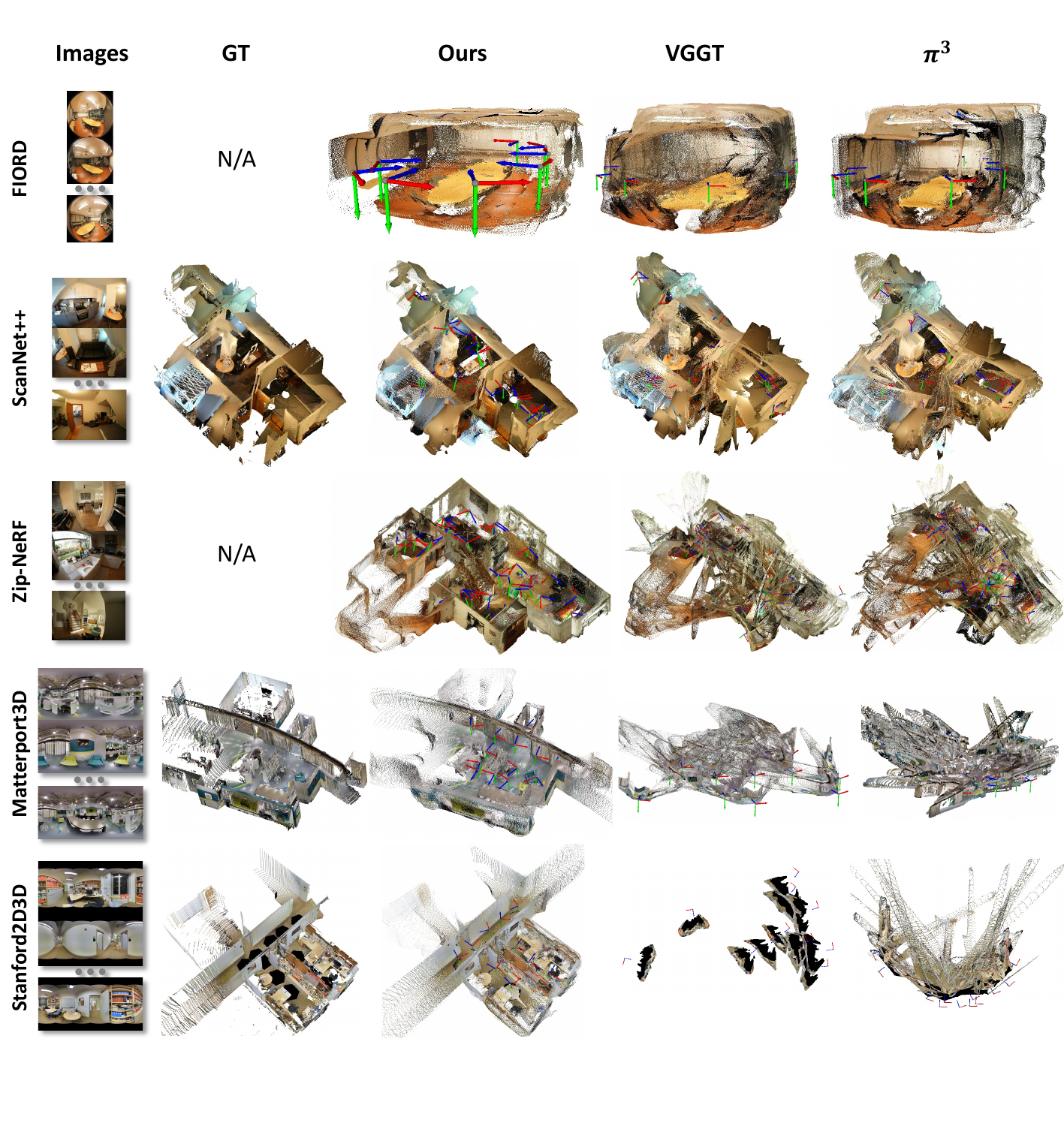}
  \vspace*{-13mm}
  \caption{
  Visualization of feed-forward 3D reconstruction on wide field-of-view images. 
  Our method shows robust performance on fisheye images from FIORD \cite{gunes2025fiord}, Zip-NeRF \cite{duckworth2023smerf}, ScanNet++ \cite{yeshwanth2023scannet++}, and 360$^\circ$ images from Matterport3D \cite{chang2017matterport3d} and Stanford2D3D \cite{armeni2017joint}.
  }
  \label{fig_result}
  \vspace*{-4mm}
\end{figure*}
\begin{table*}[t]
    \begin{center}
    \caption{\label{table:pointmap} 
    Point map estimation on wide field-of-view images. Ours denotes models trained with our composed wide-angle training data, implemented on the $\pi^3$ baseline (Ours ($\pi^3$)) and our ray-based module (Ours (Wid3R)).
    }
    \vspace*{-4mm}
    \resizebox{0.95\textwidth}{!}{
    \begin{tabular}{l cccccc cccccc cccccc}
        \toprule
        \multirow{3}{*}{Method} & \multicolumn{6}{c}{ScanNet++ \cite{yeshwanth2023scannet++}} & \multicolumn{6}{c}{Matterport3D \cite{chang2017matterport3d}} & \multicolumn{6}{c}{Stanford2D3D \cite{armeni2017joint}} 
        \\
        \cmidrule(lr){2-7}\cmidrule(lr){8-13}\cmidrule(lr){14-19}
        & \multicolumn{2}{c}{Acc. $\downarrow$} & \multicolumn{2}{c}{Comp. $\downarrow$} & \multicolumn{2}{c}{N.C. $\uparrow$} & \multicolumn{2}{c}{Acc. $\downarrow$} & \multicolumn{2}{c}{Comp. $\downarrow$} & \multicolumn{2}{c}{N.C. $\uparrow$} & \multicolumn{2}{c}{Acc. $\downarrow$} & \multicolumn{2}{c}{Comp. $\downarrow$} & \multicolumn{2}{c}{N.C. $\uparrow$}
        \\
        \cmidrule(lr){2-3}\cmidrule(lr){4-5}\cmidrule(lr){6-7}\cmidrule(lr){8-9}\cmidrule(lr){10-11}\cmidrule(lr){12-13}\cmidrule(lr){14-15}\cmidrule(lr){16-17}\cmidrule(lr){18-19}
        & Mean & Med. & Mean & Med. & Mean & Med. & Mean & Med. & Mean & Med. & Mean & Med. & Mean & Med. & Mean & Med. & Mean & Med.
        \\
        \midrule
        VGGT \cite{wang2025vggt} & 0.135  & 0.093  & 0.068 & 0.031 & 0.704 & 0.855 & 0.327 & 0.258 & 1.756 & 1.494 & 0.530 & 0.536 & 0.387 & 0.362 & 2.115 & 1.812 & 0.536 & 0.546
        \\
        $\pi^3$ \cite{wang2025pi} & 0.086 & 0.052 & 0.037 & 0.019 & 0.739 & 0.904 & 0.315 & 0.270 & 1.308 & 1.092 & 0.539 & 0.555 & 0.380 & 0.316 & 0.745 & 0.574 & 0.557 & 0.598
        \\
        Ours ($\pi^3$ \cite{wang2025pi}) & 0.027 & 0.011 & 0.014 & 0.007 & 0.781 & 0.930 & 0.161 & 0.080 & 0.142 & 0.061 & 0.610 & 0.702 & 0.211 & 0.118 & 0.273 & 0.167 & 0.596 & 0.686 \\
        Ours (Wid3R) & \textbf{0.018} & \textbf{0.008} & \textbf{0.012} & \textbf{0.006} & \textbf{0.803} & \textbf{0.951} & \textbf{0.094} & \textbf{0.049} & \textbf{0.087} & \textbf{0.035} & \textbf{0.790} & \textbf{0.928} & \textbf{0.197} & \textbf{0.099} & \textbf{0.172} & \textbf{0.101} & \textbf{0.729} & \textbf{0.868}
        \\
        \bottomrule
    \end{tabular}
    }
    \end{center}
    \vspace*{-5mm}
\end{table*}

\begin{table*}[t]
    \centering
    \caption{Quantitative comparison of monocular depth estimation performance on the Matterport3D (M) test sets. 
    (a) denotes models trained directly on the Matterport3D training set, and (b) denotes methods based on foundation models.
    Among foundation model based methods, DAC and UniK3D require training the model itself, while 360MonoDepth and RPG360 estimates depth via post optimization. Although all of these methods are designed for single view inputs, Wid3R achieves comparable performance in single view 360$^\circ$ depth estimation to state of the art methods.
    }
    \vspace*{-4mm}
    \resizebox{0.85\linewidth}{!}{
    \begin{tabular}{ll|c|cc|c|cc|ccc}
        \toprule
        & \multirow{2}{*}{Method} & \multirow{2}{*}{Backbone} & \multirow{2}{*}{360 Train.} & \multirow{2}{*}{Opti.} & \multirow{1}{*}{Dataset} & \multicolumn{2}{c|}{Lower is better} & \multicolumn{3}{c}{Higher is better} \\ 
        & & & & & Train $\rightarrow$ Test & Abs Rel & RMSE & $\delta_{1.25}$ & $\delta_{1.25^2}$ & $\delta_{1.25^3}$ \\
        \midrule
        \multirow{8}{*}{(a)} & SliceNet \cite{pintore2021slicenet} & ResNet50 \cite{he2016deep} & \textcolor{ForestGreen}{\ding{51}} & \textcolor{Red}{\ding{55}} & M $\rightarrow$ M & 0.176 & 0.613 & 0.872 &  0.948 & 0.972 \\
        & UniFuse \cite{jiang2021unifuse} & ResNet34 \cite{he2016deep} & \textcolor{ForestGreen}{\ding{51}} & \textcolor{Red}{\ding{55}} & M $\rightarrow$ M & 0.106 & 0.494 & 0.890 & 0.962 & 0.983\\
        & EGFormer \cite{yun2023egformer} & Transformer \cite{dosovitskiy2020image} & \textcolor{ForestGreen}{\ding{51}} & \textcolor{Red}{\ding{55}} & M $\rightarrow$ M & 0.147 & 0.603 & 0.816 & 0.939 & 0.974 \\
        & ACDNet \cite{zhuang2022acdnet} & ResNet50 \cite{he2016deep} & \textcolor{ForestGreen}{\ding{51}} & \textcolor{Red}{\ding{55}} & M $\rightarrow$ M & 0.101 & 0.463 & 0.900 & 0.968 & 0.988 \\
        & BiFuse++ \cite{wang2022bifuse++} & ResNet34 \cite{he2016deep} & \textcolor{ForestGreen}{\ding{51}} & \textcolor{Red}{\ding{55}} & M $\rightarrow$ M & 0.112 & 0.485 & 0.881 & 0.966 & 0.987\\
        & HRDFuse \cite{ai2023hrdfuse} & ResNet34 \cite{he2016deep} & \textcolor{ForestGreen}{\ding{51}} & \textcolor{Red}{\ding{55}} & M $\rightarrow$ M & 0.117 & 0.503 & 0.867 & 0.962 & 0.985 \\
        & Elite360D \cite{ai2024elite360d} & EfficientNet-B5 \cite{tan2019efficientnet} & \textcolor{ForestGreen}{\ding{51}} & \textcolor{Red}{\ding{55}} & M $\rightarrow$ M & 0.105 & \bf{0.452} & 0.899 & 0.971 & \bf{0.991} \\
        & Depth Anywhere \cite{wang2024depth} & BiFuse$++$ \cite{wang2022bifuse++} & \textcolor{ForestGreen}{\ding{51}} & \textcolor{Red}{\ding{55}} & M $\rightarrow$ M & \textbf{0.085} & - & 0.917 & \bf{0.976} & \bf{0.991} \\
        \midrule
        \multirow{5}{*}{(b)} & 360MonoDepth \cite{rey2022360monodepth} & MiDaS v2 \cite{ranftl2021vision} & \textcolor{Red}{\ding{55}} & \textcolor{ForestGreen}{\ding{51}} & \includegraphics[height=1em]{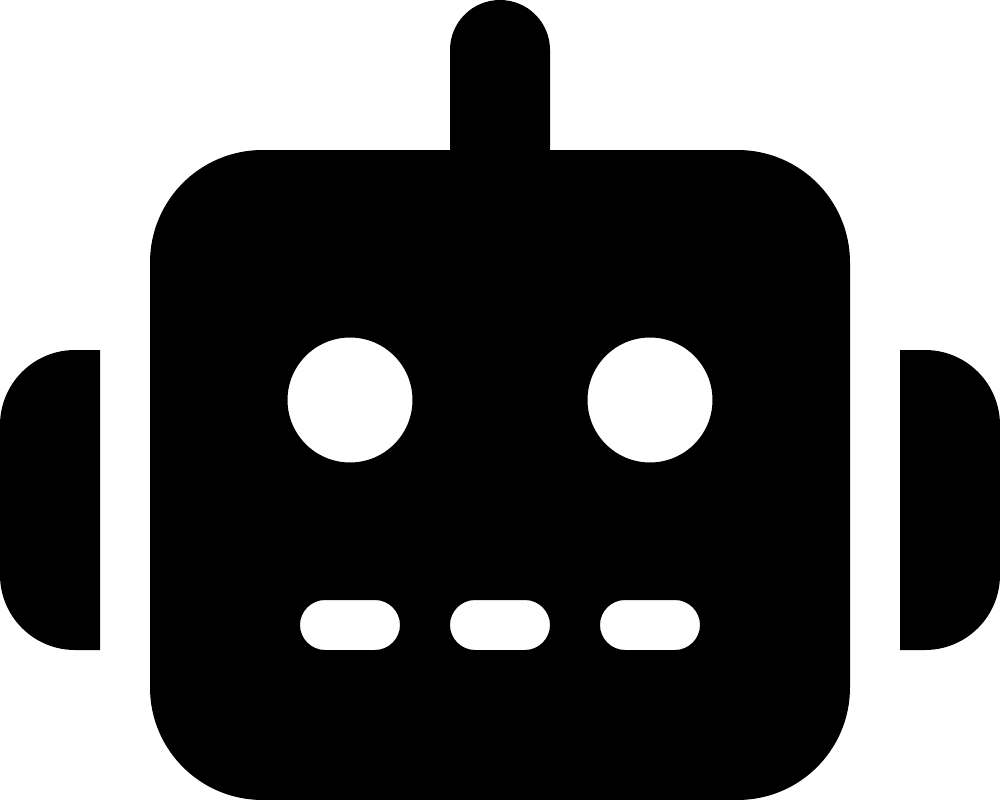} $\rightarrow$ M & 0.264 & 0.916 & 0.612 & 0.854 & 0.941 \\
        & DAC \cite{guo2025depth} & ResNet101 \cite{he2016deep} & \textcolor{ForestGreen}{\ding{51}} & \textcolor{Red}{\ding{55}} & \includegraphics[height=1em]{figures/robot-solid.pdf} $\rightarrow$ M &  0.156 & 0.619 & 0.773 & 0.956 & 0.982 \\
        & UniK3D \cite{piccinelli2025unik3d} & DINOv2 \cite{oquab2023dinov2} & \textcolor{ForestGreen}{\ding{51}} & \textcolor{Red}{\ding{55}} & \includegraphics[height=1em]{figures/robot-solid.pdf} $\rightarrow$ M & 0.315 & 0.649 & 0.858 & 0.957 & 0.977 \\
        & RPG360 \cite{jung2025rpg360} & Metric3D v2 \cite{hu2024metric3d} & \textcolor{Red}{\ding{55}} & \textcolor{ForestGreen}{\ding{51}} & \includegraphics[height=1em]{figures/robot-solid.pdf} $\rightarrow$ M & 0.203 & 0.667 & 0.859 & 0.953 & 0.977  \\
        & Ours (Wid3R) & DINOv2 \cite{oquab2023dinov2} & \textcolor{ForestGreen}{\ding{51}} & \textcolor{Red}{\ding{55}} & \includegraphics[height=1em]{figures/robot-solid.pdf} $\rightarrow$ M & 0.227 & 0.562 & \textbf{0.948} & \textbf{0.976} & 0.984 \\
        \bottomrule
    \end{tabular}
    }
    \label{table:360depth}
    \vspace*{-3mm}
\end{table*}
\begin{table*}[t]
    \begin{center}
    \caption{\label{table:ablation} 
    Ablation study on Stanford2D3D \cite{armeni2017joint} under the 360$^\circ$ setting, which is not included in the training data. 
    We analyze the effects of camera model tokens, training data composition (pinhole, fisheye, 360), and the number of frames per batch using point map estimation metrics.
    Notably, this study demonstrates that training with diverse camera models improves performance on 360$^\circ$ images.
    }
    \vspace*{-4mm}
    \resizebox{0.99\textwidth}{!}{
    \begin{tabular}{c 
    @{\hspace{10pt}} c
    @{\hspace{10pt}} c
    @{\hspace{10pt}} c
    @{\hspace{10pt}} c
    @{\hspace{10pt}} c
    @{\hspace{10pt}}cccccc 
    @{\hspace{10pt}}cccccc
    }
        \toprule
        \multirow{2}{*}{Study} & \multirow{2}{*}{\makecell{Frames\\per batch}} & \multirow{2}{*}{\makecell{Camera Token\\Condition}} & \multicolumn{3}{c}{Training Data} & \multicolumn{2}{c}{Acc. $\downarrow$} & \multicolumn{2}{c}{Comp. $\downarrow$} & \multicolumn{2}{c}{N.C. $\uparrow$} & \multicolumn{2}{c}{Acc. std. $\downarrow$} & \multicolumn{2}{c}{Comp. std. $\downarrow$} & \multicolumn{2}{c}{N.C. std. $\downarrow$} 
        \\
        \cmidrule(lr){4-6}\cmidrule(lr){7-8}\cmidrule(lr){9-10}\cmidrule(lr){11-12}\cmidrule(lr){13-14}\cmidrule(lr){15-16}\cmidrule(lr){17-18}        
        & & & Pinhole & Fisheye & Sphere & Mean & Med. & Mean & Med. & Mean & Med. & Mean & Med. & Mean & Med. & Mean & Med.
        \\
        \midrule
        \multirow{4}{*}{(a)} & 2 $\sim$ 12 & \textcolor{ForestGreen}{\ding{51}} & \textcolor{Red}{\ding{55}} & \textcolor{Red}{\ding{55}} & \textcolor{ForestGreen}{\ding{51}} & 0.220 & 0.142 & 0.355 & 0.235 & 0.710 & 0.827 & 0.187 &	0.135 &	0.416 &	0.311 &	0.130 &	0.197
        \\
        & 2 $\sim$ 12 & \textcolor{ForestGreen}{\ding{51}} & \textcolor{ForestGreen}{\ding{51}} & \textcolor{Red}{\ding{55}} & \textcolor{ForestGreen}{\ding{51}} & 0.254 & 0.128 & 0.166 & 0.108 & 0.698 & 0.839 & 0.099 & 0.107 & 0.243 & 0.206 & 0.099 & 0.149
        \\
        & 2 $\sim$ 12 & \textcolor{ForestGreen}{\ding{51}} & \textcolor{Red}{\ding{55}} & \textcolor{ForestGreen}{\ding{51}} & \textcolor{ForestGreen}{\ding{51}} & 0.593 & 0.126 & \textbf{0.120} & \textbf{0.056} & 0.686 & 0.822 & 0.101 &	0.109 &	\textbf{0.136} & \textbf{0.071} & 0.100 &	0.173
        \\
        & 2 $\sim$ 12 & \textcolor{ForestGreen}{\ding{51}} & \textcolor{ForestGreen}{\ding{51}} & \textcolor{ForestGreen}{\ding{51}} & \textcolor{ForestGreen}{\ding{51}} & 0.248 & 0.131 & 0.146 & 0.074 & 0.706 & 0.842 & \textbf{0.089} & 0.103 & 0.164 & 0.090 & \textbf{0.084} & \textbf{0.134}
        \\
        \midrule
        (b) & 2 $\sim$ 12& \textcolor{Red}{\ding{55}} & \textcolor{ForestGreen}{\ding{51}} & \textcolor{ForestGreen}{\ding{51}} & \textcolor{ForestGreen}{\ding{51}} & 0.284 & 0.145 & 0.152 & 0.089 & 0.680 & 0.803 & 0.103 & 0.095 & 0.178 & 0.147 & 0.091 & 0.152
        \\
        \midrule
        (c) & 2 $\sim$ 24 & \textcolor{ForestGreen}{\ding{51}} & \textcolor{ForestGreen}{\ding{51}} & \textcolor{ForestGreen}{\ding{51}} & \textcolor{ForestGreen}{\ding{51}} & \textbf{0.197} & \textbf{0.099} & 0.172 & 0.101 & \textbf{0.729} & \textbf{0.868} & 0.113 & \textbf{0.089} & 0.257 & 0.216 &	0.095 &	0.143
        \\
        \bottomrule
    \end{tabular}
    }
    \end{center}
    \vspace*{-6mm}
\end{table*}

\subsection{Camera Pose Estimation}\label{sec:pose_estimation}
\noindent\textbf{Pose Evaluation}
We perform zero-shot camera pose evaluation on wide field-of-view cameras, including fisheye cameras on FIORD \cite{gunes2025fiord} and Zip-NeRF \cite{duckworth2023smerf}, and 360$^\circ$ cameras on Stanford2D3D \cite{armeni2017joint}.
For Stanford2D3D, we randomly select 10-30 images from each scene in \textit{area\_5a} and \textit{area\_5b}, repeating this sampling process 10 times per scene to construct a total of 20 cases.
For FIORD, we randomly select an initial frame and then sample frames at an interval of five, resulting in 4–24 images per sequence from the \textit{Kitchen\_In}, \textit{meetingroom}, and \textit{parakennus} scenes. 
For Zip-NeRF, we follow the official test split provided in \cite{pataki2025mp}.
Following \cite{wang2025pi}, we evaluate camera pose accuracy using Relative Rotation Accuracy (RRA) and Relative Translation Accuracy (RTA) at predefined thresholds. 
We further report the Area Under the Curve (AUC) of the $\min(\mathrm{RRA}, \mathrm{RTA})$ threshold curve as a unified performance metric, as summarized in Table~\ref{table:pose_angle}.
In addition, we report Absolute Trajectory Error (ATE), Relative Pose Error for translation (RPE trans), and Relative Pose Error for rotation (RPE rot) in Table~\ref{table:pose_distance}.
Compared to other state-of-the-art methods \cite{wang2025vggt,wang2025pi}, Wid3R significantly outperforms prior approaches in camera pose estimation, demonstrating superior accuracy and robustness for wide field-of-view camera settings.

\noindent\textbf{Large-scale Localization}
Following \cite{jung2025im360}, we report AUC scores computed from the maximum of the relative rotation and translation errors across all image pairs. 
We compare our results with state-of-the-art methods in Table \ref{table:vl_mp3d} on three scenes from the Matterport3D dataset.
Existing localization approaches for 360$^\circ$ imagery predominantly rely on structure-from-motion (SfM) pipelines involving triangulation \cite{hartley2003multiple} and bundle adjustment \cite{triggs1999bundle}. 
While IM360 \cite{jung2025im360} achieves accurate localization performance through SfM, it requires dense matching EDM \cite{jung2025edm} across all image pairs, which introduces significant computational bottlenecks and leads to long processing times.
In contrast, Wid3R adopts a feed-forward formulation and can produce 3D maps within three seconds even in large-scale indoor environments, demonstrating both computational efficiency and scalability.
Figure \ref{fig_sfm} demonstrates a qualitative comparison of the visual results.


\vspace*{-1.5mm}
\subsection{Point Map Estimation}
\vspace*{-1.5mm}
To evaluate point map estimation under wide field-of-view cameras, we conduct experiments on the ScanNet++ \cite{yeshwanth2023scannet++}, Matterport3D \cite{chang2017matterport3d}, and Stanford2D3D \cite{armeni2017joint} datasets.
For ScanNet++, we use the official \textit{nvs\_sem\_val} split and randomly sample 20–30 images per scene from 50 scenes. For Matterport3D, we follow the official test split consisting of 18 scenes and sample images to ensure sufficient visual overlap using the probability matrix described in Section ~\ref{sec:Training}. For Stanford2D3D, we use the same test cases as those described in Section~\ref{sec:pose_estimation}.
For point cloud alignment, we sequentially apply Umeyama alignment \cite{umeyama2002least}, optimal point alignment \cite{wang2025moge}, followed by Iterative Closest Point.
Following \cite{wang2025continuous,wang2025pi}, we report Accuracy (Acc.), Completion (Comp.), and Normal Consistency (N.C.) as evaluation metrics.
Table~\ref{table:pointmap} demonstrates that Wid3R achieves superior performance, particularly under 360$^\circ$ camera settings. 
We further compare Wid3R with Ours ($\pi^3$ \cite{wang2025pi}), which is designed based on the $\pi^3$ architecture without the ray representation and camera model tokens, showing that our method provides clear advantages.
Qualitative results are presented in Fig.~\ref{fig_result}, illustrating the robustness of Wid3R across diverse camera types. 
Notably, while prior methods exhibit performance degradation on distorted imagery, our approach enables accurate estimation even for 360$^\circ$ images.

\subsection{Monocular 360 Depth Estimation}
We conduct monocular panoramic depth estimation evaluation on Matterport3D \cite{chang2017matterport3d}, as in \cite{eigen2014depth,jung2025im360}. 
Table \ref{table:360depth} presents the quantitative results. 
Wid3R achieves comparable or superior performance to recent monocular 360$^\circ$ depth estimation methods, even though it is trained in a multi-view setting and uses not only 360$^\circ$ cameras but also pinhole and fisheye cameras.

\subsection{Ablation Study}
To validate the effectiveness of our proposed method, we conduct an ablation study using point map estimation metrics on Stanford2D3D \cite{armeni2017joint}, with results reported in Table~\ref{table:ablation}. 
We first analyze the impact of training data composition. Compared to other camera types, 360$^\circ$ datasets contain significantly fewer geometry-labeled samples, accounting for less than $1\%$ of the total training data. 
To address this imbalance, prior work has proposed generating pseudo-labels using large-scale perspective models \cite{wang2024depth} or applying optimization-based post-processing methods \cite{rey2022360monodepth,jung2025rpg360}. 
In contrast, we investigate whether incorporating data from other camera models can improve performance on 360$^\circ$ cameras. 
As shown in Table~\ref{table:ablation}(a), jointly training with additional camera model datasets improves geometry estimation performance on 360$^\circ$ data while reducing performance variance, compared to training solely on 360$^\circ$ images. 
This indicates that, while camera model tokens provide explicit conditioning, exposing the shared network parameters to more diverse training data mitigates overfitting caused by limited 360$^\circ$ supervision and leads to more stable and robust learning.
We further observe in Table~\ref{table:ablation}(b) that incorporating camera model tokens as prompts consistently improves performance over models trained without them, highlighting the benefit of explicitly encoding camera configuration. 
Finally, we examine the effect of the maximum number of frames per batch in Table~\ref{table:ablation}(c). 
While hardware constraints limit the maximum number to 12 frames on an RTX A6000 GPU, up to 24 frames can be used on an H100 GPU. 
Increasing the number of multi-view images per batch improves Acc. and N.C., indicating that larger numbers of frames during training enable the model to learn richer multi-view attention patterns and yield more accurate geometry estimation.

\section{Conclusion, Limitations, and Future Work}
We introduce a novel feed-forward 3D reconstruction method designed for wide field-of-view camera models. To address the inherent distortions of large field-of-view imagery, our approach incorporates ray-based geometry representation modules and camera model tokens as explicit conditioning prompts. We further demonstrate that our method achieves superior accuracy in monocular depth estimation, visual localization, and 3D reconstruction, with particularly strong performance on 360$^\circ$ datasets, which are challenging for large-scale model training due to the limited availability of labeled benchmarks.
One limitation of our approach is that, while it supports wide field-of-view inputs and large-scale environments, it does not explicitly model dynamic scene elements. As future work, we plan to extend the framework by incorporating datasets with dynamic content and designing additional modules to handle scene dynamics.

\clearpage  

%
%
\bibliographystyle{splncs04}
\bibliography{main}

\title{Supplementary Material of Wid3R:\\ Wide Field-of-View 3D Reconstruction via Camera Model Conditioning} 

\titlerunning{Wid3R}

\author{Dongki Jung\inst{1,2} \and  Jaehoon Choi\inst{1} \and Adil Qureshi\inst{1} \and Somi Jeong\inst{2} \and\\ Dinesh Manocha\inst{1} \and Suyong Yeon\inst{2}
}

\authorrunning{Jung et al.}

\institute{$^{1}$University of Maryland, College Park and $^{2}$NAVER LABS}

{
\maketitle
\begin{figure*}[ht]
  \centering
  \includegraphics[width=1.0\textwidth]{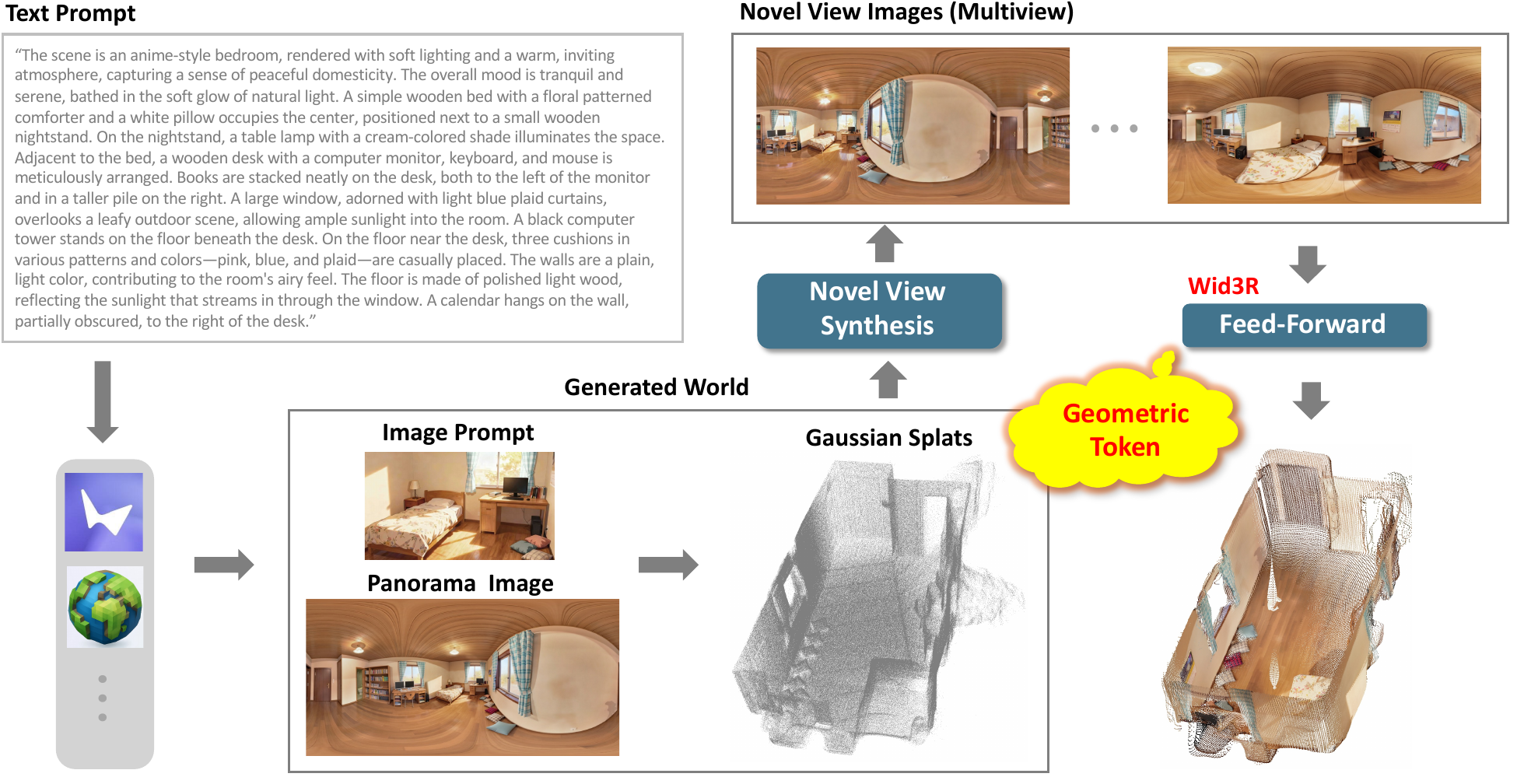}
  \caption{
  Overview of the applicability of Wid3R in panorama-based scene generation. The text describes the generation of panorama images and 3D information, such as panoramic depth and Gaussian splats \cite{marble2025, worldgen}. We tested Wid3R on the synthesized images, demonstrating that it can potentially enhance scene generation by providing geometric token information, improving consistency across generated images.
  }
  \label{supp:fig_marble}
  \vspace*{-8mm}
\end{figure*}
}

\section{Potential for Panorama-Based Scene Generation}
Recent works such as WorldGen \cite{worldgen}, Marble \cite{marble2025}, and One2Scene \cite{wang2026one2scene} generate panoramic images using panorama generation models \cite{team2025hunyuanworld,dit360,wu2023ipoldm,zhang2024taming,ni2025makes}, and subsequently reconstruct 3D environments from these panoramas to enable novel view synthesis. 
Such panorama-based generation pipelines have also been explored \footnote{\url{https://developer.nvidia.com/blog/simulate-robotic-environments-faster-with-nvidia-isaac-sim-and-world-labs-marble/}} for creating large-scale synthetic environments for robotic policy training \cite{li2023panogen,wang2025panogen++}.
We investigate whether Wid3R can be applied to these panorama generation frameworks, as shown in Fig. \ref{supp:fig_marble}.
Since Wid3R can extract geometric token information from panoramic images, these tokens could potentially provide additional guidance during the generative process or enforce multi-view geometric consistency across generated scenes. 
While we do not integrate Wid3R into the generation pipeline in this work, our preliminary exploration suggests that Wid3R could serve as a useful geometric prior for panorama-based scene generation, which we leave as future work.

\section{Data Configuration}
\begin{table*}[t]
    \centering
    \begin{minipage}{0.45\textwidth}
    \caption{Training Data}
    \label{supp:table_traindata}
    \vspace*{-4mm}
    \resizebox{\textwidth}{!}{
    \begin{tabular}{l|@{\hspace{4mm}}c@{\hspace{4mm}}c@{\hspace{4mm}}c}
    \hline
    \textbf{Name} & \textbf{Scene} & \textbf{Images} & \textbf{Camera} \\
    \hline
    Hypersim \cite{roberts2021hypersim}
     & 365  & 59543  & Pinhole \\
    EDEN \cite{le2021eden}         & 630  & 316385 & Pinhole \\
    vKITTI \cite{cabon2020virtual}       & 50   & 42520  & Pinhole \\
    ASE \cite{engel2023project}          & 1000 & 593051 & Fisheye \\
    ScanNet++ \cite{yeshwanth2023scannet++}    & 856  & 962368 & Fisheye \\
    KITTI360 \cite{liao2022kitti}     & 72   & 141768 & Fisheye \\
    TartanAirV2 \cite{wang2020tartanair}  & 65   & 80467  & Sphere     \\
    Matterport3D \cite{chang2017matterport3d} & 61   & 7829   & Sphere     \\
    360Loc \cite{huang2024360loc}       & 4    & 2244   & Sphere     \\
    \hline
    \end{tabular}
    }
    \end{minipage}%
    \hfill%
    \begin{minipage}{0.45\textwidth}
    \caption{Evaluation Data}
    \resizebox{\textwidth}{!}{
    \begin{tabular}{l|@{\hspace{4mm}}c@{\hspace{4mm}}c@{\hspace{4mm}}c}
    \hline
    \textbf{Name}     & \textbf{Scene} & \textbf{Images} & \textbf{Camera} \\
    \hline
    FIORD  \cite{gunes2025fiord}          & 3    & 253    & Fisheye \\
    ScanNet++ \cite{yeshwanth2023scannet++}       & 50   & 1244   & Fisheye \\
    Zip-NeRF \cite{duckworth2023smerf}        & 4    & 815    & Fisheye \\
    Matterport3D \cite{chang2017matterport3d}    & 18   & 358    & Sphere     \\
    Stanford2D3D \cite{armeni2017joint}    & 2    & 239    & Sphere     \\
    \hline
    \end{tabular}
    }
    \end{minipage}
    \vspace*{-4mm}
\end{table*}
The ASE dataset \cite{engel2023project} has a very large total size, so we only utilized 1000 scenes. 
For KITTI-360 \cite{liao2022kitti}, we divide the data into sub-scenes for each of the 1000 sequences. 
As shown in Table \ref{supp:table_traindata}, the total number of spherical images is significantly smaller compared to other image types. 
To ensure performance stability, we implemented uniform sampling probabilities for each training data. 
Additionally, we applied rotation augmentation to the 360$^\circ$ images during training to prevent overfitting in the 360$^\circ$ training datasets.

\begin{algorithm}
\centering
\caption{Softmin Probability Calculation}
\begin{algorithmic}[1]
\State \textbf{Input:} $C$: $N \times N$ cost matrix  \Comment{$N$ is the number of images}
\State \textbf{Output:} $P$: $N \times N$ probability matrix
\State Set diagonal of $C$ to infinity
\For{each row $i$ in $C$}
    \State Find minimum value in row $i$
    \State Subtract the minimum value from each element in row $i$
\EndFor
\State Set $\tau$ to the median of each row's values
\For{each element $i, j$ in $C$}
    \State Calculate weight $w[i, j] = \exp(-C[i, j] / \tau)$
\EndFor
\State $\Sigma$ $\gets$ Calculate row sums of weights  \Comment{$\Sigma \in \mathbb{R}^{N \times 1}$}
\State $P$ $\gets$ Normalize weights $w$ by dividing by $\Sigma$
\State \textbf{Return:} $P$
\end{algorithmic}
\label{supp:algorithm}
\end{algorithm}


\section{Data Preprocessing}
For training the Wid3R model, we perform batch selection by first gathering a set of images within a single scene. 
Subsequently, we calculate the distances between the various camera positions, where the distance is defined as the difference between camera positions in the world coordinate system. 
Using these distances, we construct a Softmin probability map, as outlined in Algorithm \ref{supp:algorithm}. 
This probability matrix indicates the likelihood of selecting images that are spatially close, thereby prioritizing those with higher proximity for inclusion in the same batch during training. 
In our experiments, we utilized wide field-of-view images, and our findings suggest that this straightforward image selection method, based on camera translation, effectively generates valid image overlaps. 
These overlaps, in turn, contribute to the training of our feedforward network, ensuring meaningful spatial correlations for improved performance.

\section{Network Architecture}
\begin{table*}[t]
    \begin{center}
    \caption{\label{supp:table_pointmap} 
    Performance comparison of different attention methods applied in Wid3R.
    }
    \vspace*{-2mm}
    \resizebox{0.95\textwidth}{!}{
    \begin{tabular}{l cccccc cccccc cccccc}
        \toprule
        \multirow{3}{*}{Method} & \multicolumn{6}{c}{ScanNet++ \cite{yeshwanth2023scannet++}} & \multicolumn{6}{c}{Matterport3D \cite{chang2017matterport3d}} & \multicolumn{6}{c}{Stanford2D3D \cite{armeni2017joint}} 
        \\
        \cmidrule(lr){2-7}\cmidrule(lr){8-13}\cmidrule(lr){14-19}
        & \multicolumn{2}{c}{Acc. $\downarrow$} & \multicolumn{2}{c}{Comp. $\downarrow$} & \multicolumn{2}{c}{N.C. $\uparrow$} & \multicolumn{2}{c}{Acc. $\downarrow$} & \multicolumn{2}{c}{Comp. $\downarrow$} & \multicolumn{2}{c}{N.C. $\uparrow$} & \multicolumn{2}{c}{Acc. $\downarrow$} & \multicolumn{2}{c}{Comp. $\downarrow$} & \multicolumn{2}{c}{N.C. $\uparrow$}
        \\
        \cmidrule(lr){2-3}\cmidrule(lr){4-5}\cmidrule(lr){6-7}\cmidrule(lr){8-9}\cmidrule(lr){10-11}\cmidrule(lr){12-13}\cmidrule(lr){14-15}\cmidrule(lr){16-17}\cmidrule(lr){18-19}
        & Mean & Med. & Mean & Med. & Mean & Med. & Mean & Med. & Mean & Med. & Mean & Med. & Mean & Med. & Mean & Med. & Mean & Med.
        \\
        \midrule
        Ours w/ Cross-attn & 0.022 & 0.011 & 0.013 & 0.007 & 0.782 & 0.938 & 0.112 & 0.055 & 0.093 & 0.037 & 0.753 & 0.907 & 0.273 & 0.167 & 0.211 & 0.126 & 0.660 & 0.774  \\
        Ours w/ Self-attn & \textbf{0.018} & \textbf{0.008} & \textbf{0.012} & \textbf{0.006} & \textbf{0.803} & \textbf{0.951} & \textbf{0.094} & \textbf{0.049} & \textbf{0.087} & \textbf{0.035} & \textbf{0.790} & \textbf{0.928} & \textbf{0.197} & \textbf{0.099} & \textbf{0.172} & \textbf{0.101} & \textbf{0.729} & \textbf{0.868}
        \\
        \bottomrule
    \end{tabular}
    }
    \end{center}
    \vspace*{-4mm}
\end{table*}

\begin{figure*}[t]
  \centering
  \includegraphics[width=1.0\textwidth]{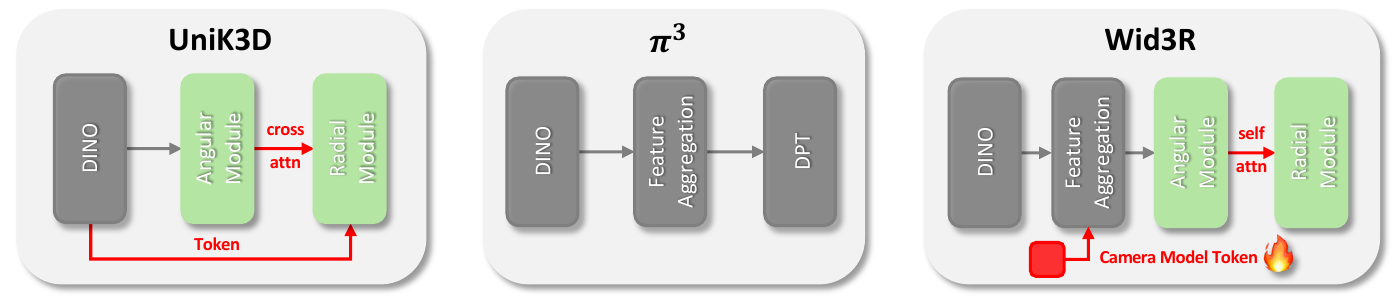}
  \caption{
  Building upon UniK3D \cite{piccinelli2025unik3d} and $\pi^3$ \cite{wang2025pi}, Wid3R incorporates feature extraction along with Angular and Radial modules.
  The key distinction from these previous works is the inclusion of a camera model token to represent camera modeling and the use of self-attention within the radial module.
  }
  \label{supp:fig_network}
\end{figure*}
Inspired by UniK3D \cite{piccinelli2025unik3d}, we introduce Radial and Angular modules as headers in Wid3R, modifying the original architecture of $\pi^3$ \cite{wang2025pi}. 
However, our design diverges from UniK3D in several key aspects. 
While UniK3D extracts token information from DINO \cite{oquab2023dinov2} to drive angular module prediction, and utilizes cross-attention between the features from the angular and radial modules, Wid3R takes a different approach. 
Instead of relying on DINO, Wid3R employs the camera model token, which is explicitly defined to facilitate the loading of pretrained weights from $\pi^3$ \cite{piccinelli2025unik3d}, thereby accelerating training convergence. 
This choice, however, leads to performance degradation when cross-attention is applied between the features of the radial and angular modules. 
To address this, we incorporate self-attention within the radial module. A performance comparison of these architectures is presented in Table \ref{supp:table_pointmap}.
Figure \ref{supp:fig_network} illustrates the key difference between previous methods and our approach.

\section{Ray Representation}
\begin{figure*}[t]
  \centering
  \begin{minipage}{0.6\textwidth}
    \centering
    \includegraphics[width=\textwidth]{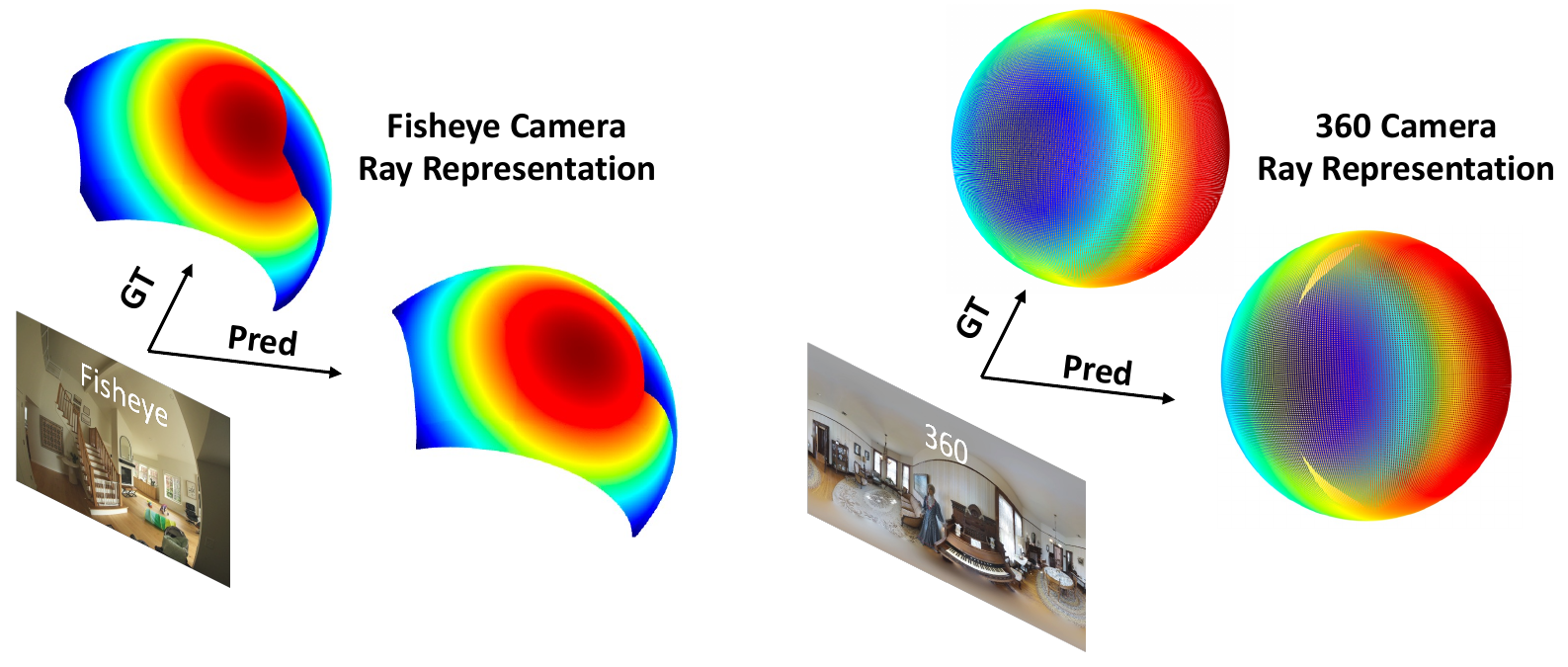}
  \end{minipage}%
  \begin{minipage}{0.35\textwidth}
    \centering
    \caption{
      The visualization of ray prediction from the Angular module, where rays are predicted using spherical harmonics.
    }
    \label{supp:fig_ray}
  \end{minipage}
\end{figure*}

Given an input image along with a camera model token, which serves as a prompt for camera-model conditioning, the Angular Module of Wid3R predicts the ray direction for each pixel. 
The token indicates the underlying projection model, e.g., a fisheye token for Zip-NeRF \cite{duckworth2023smerf} or a 360$^\circ$ token for Matterport3D \cite{chang2017matterport3d}, as shown in Fig.~\ref{supp:fig_ray}.
Subsequently, the Radial Module estimates the corresponding radial distance along the predicted ray. 
The final 3D point is obtained by scaling the predicted ray direction with the estimated radial distance, thereby determining the spatial location of the point in 3D space.

\section{Cubemap Transformation}
\begin{figure*}[t]
  \centering
  \includegraphics[width=1.0\textwidth]{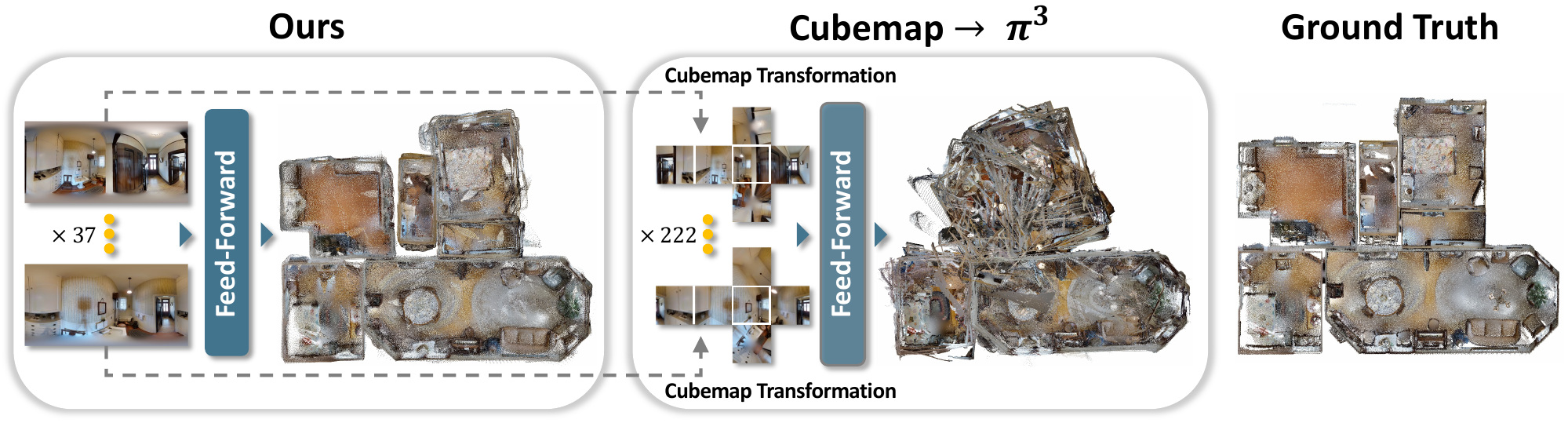}
  \caption{
  Although cubemap transformation can be applied to distorted images for estimation using a feed-forward model trained on perspective images, it results in decreased efficiency and performance.
  }
  \label{supp:fig_cubemap}
\end{figure*}
One possible approach for handling wide field-of-view images in 3D reconstruction is to first undistort the inputs and then apply models trained on perspective-based foundation models, such as $\pi^3$ \cite{wang2025pi}. 
However, this strategy is neither computationally efficient nor scalable. 
For example, a single Matterport3D scene contains 37 spherical images, which can be converted into 222 cubemap images after undistortion. 
As a result, the model must process a substantially larger number of images, leading to increased computational cost and processing time.
Moreover, for fisheye cameras, accurate undistortion requires knowledge of the camera’s distortion parameters. 
Obtaining and applying these parameters introduces additional preprocessing steps and system complexity. 
Consequently, such undistortion-based pipelines become impractical when scaling to large datasets or diverse camera types.

\section{Additional Experimental Result}
\begin{figure*}[t]
  \centering
  \includegraphics[width=1.0\textwidth]{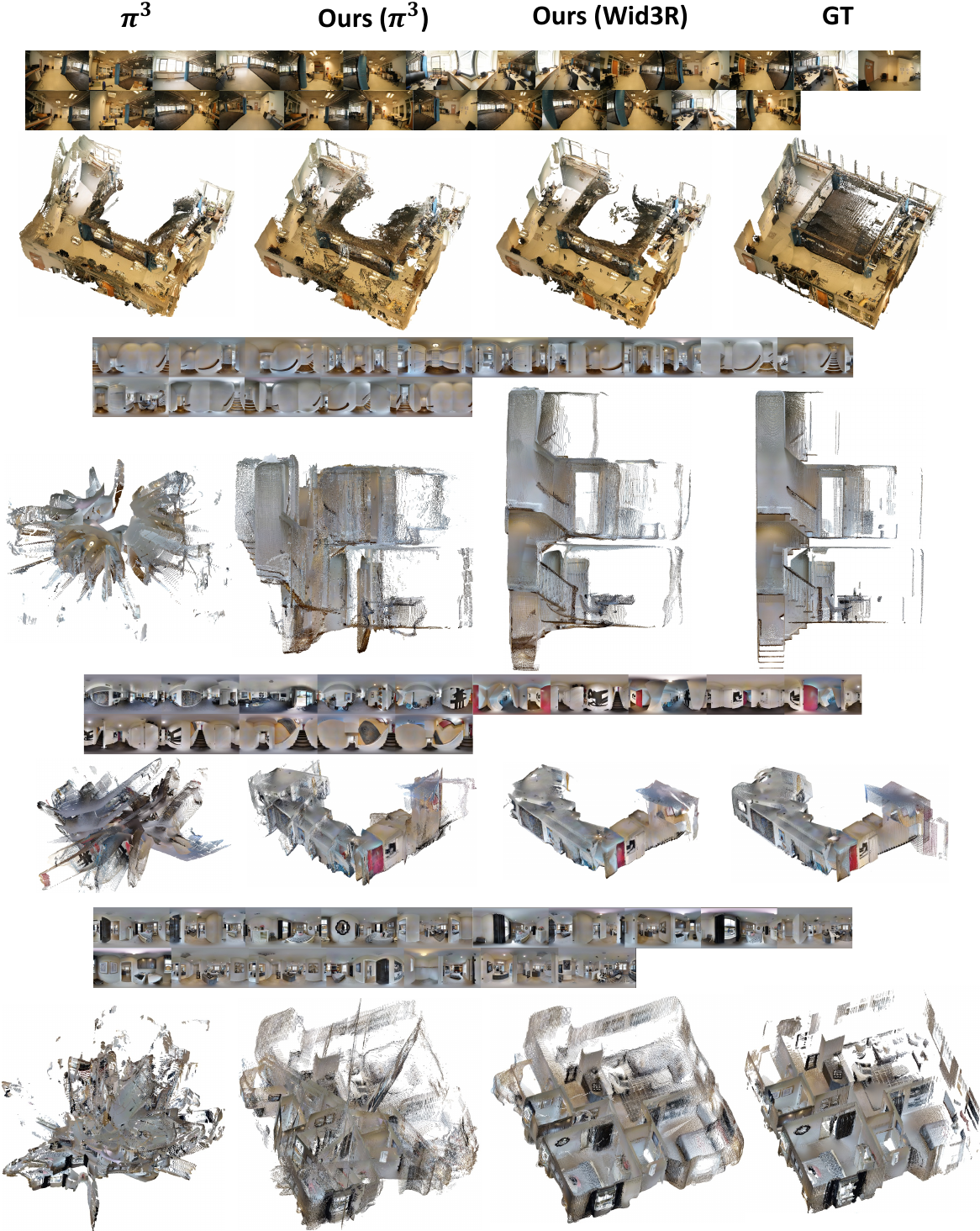}
  \caption{
  Additional visualization results of Wid3R compared with other methods.
  }
  \label{supp:fig_result1}
\end{figure*}

\begin{figure*}[t]
  \centering
  \includegraphics[width=1.0\textwidth]{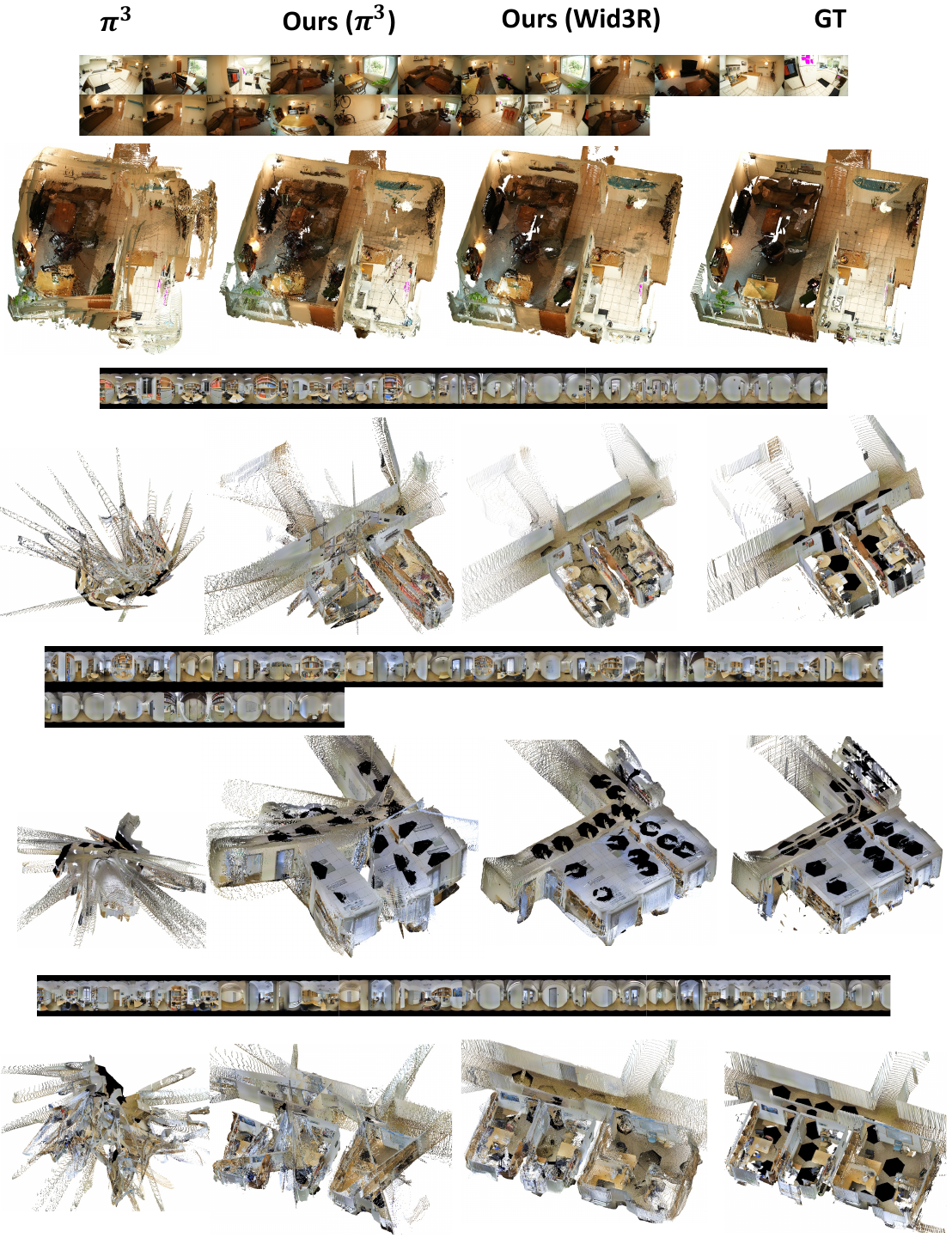}
  \caption{
  Additional visualization results of Wid3R compared with other methods.
  }
  \label{supp:fig_result2}
\end{figure*}

Figures \ref{supp:fig_result1} and \ref{supp:fig_result2} present additional qualitative results of Wid3R. 
In each row, the upper part shows the multiple input images, while the lower part visualizes the estimated 3D points. 
The first row corresponds to a fisheye camera setup, whereas the remaining rows correspond to 360$^\circ$ cameras. 
Scenes captured with 360$^\circ$ cameras often contain significant occlusions and multi-level structures, making accurate 3D prediction particularly challenging. 
Nevertheless, Wid3R produces consistent and accurate reconstructions in these scenarios. 
The baseline labeled Ours ($\pi^3$) denotes the $\pi^3$ network fine-tuned with our training dataset configuration, which still exhibits noticeable artifacts in its predictions.

\clearpage  

%
%
\end{document}